\documentclass[twocolumn]{next++}

\usepackage[toc,page,header]{appendix}
\usepackage{minitoc}
\usepackage{amsmath,amssymb}
\usepackage{bm}
\usepackage{enumitem}
\usepackage{url}
\usepackage{graphicx}
\usepackage{booktabs}
\usepackage{multirow}
\usepackage{tabularx}
\usepackage{array}
\usepackage[table]{xcolor}
\usepackage[font=small]{caption}
\usepackage{arydshln}
\newcolumntype{Y}{>{\raggedright\arraybackslash}X}
\newcolumntype{L}[1]{>{\hsize=#1\hsize
  \linewidth=\hsize\raggedright\arraybackslash}X}
\title{SuperAI: Merging Capabilities across Heterogeneous Foundation Models}

\author[1,2]{Jingtong Dou*}
\author[1,2]{Chuancheng Shi*}
\author[1]{Wenhua Wu}
\author[1]{Wei Bao}
\author[2]{Fei Shen}
\author[1]{Zhiyong Wang}
\author[2]{Tat-Seng Chua}

\affiliation[1]{University of Sydney}
\affiliation[2]{National University of Singapore, NExT++ Research Centre}

\abstract{
Enabling the reuse and composition of knowledge and capabilities distributed across foundation models is an important step toward building scalable artificial intelligence (AI) systems.
However, architectural differences and incompatible representations make it difficult to transfer and integrate these capabilities through a unified cross-model interface.
To address this challenge, we propose SuperAI, a cross-architecture multi-model fusion framework that integrates complementary capabilities from multiple source models into a host model through orthogonal alignment of low-dimensional task-relevant representations.
Specifically, we first extract task-relevant representations from the source models and the host model using shared anchor samples, and apply principal component analysis (PCA) to construct low-dimensional functional subspaces with the same dimensionality.
Next, we use orthogonal Procrustes mappings to align the source and host coordinate systems, establishing cross-model correspondences while preserving the internal geometry of each source subspace.
Finally, we train a lightweight residual multilayer perceptron (MLP) with joint supervision on task outputs and source-model increments, learning input-dependent residual increments that are injected into the host model for multi-model capability fusion.
Experiments on three host models across 15 capability domains show that SuperAI achieves higher average domain scores than the evaluated cross-architecture fusion methods and better preserves general capabilities as the number of source models increases.
Ablation studies further support the effectiveness of orthogonal alignment, nonlinear residual mapping, and joint supervision.
\vspace{-0.1cm}
}

\checkdata[Resources]{
  \href{https://github.com/jocelyndou/SuperAI_Merge.git}{\raisebox{-1ex}{\includegraphics[height=4.2ex]{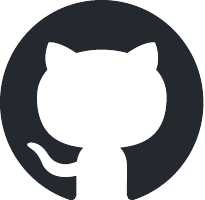}}~Code}\hspace{1em}
  \href{mailto:your.email@sydney.edu.au}
  {\texttt{jdou0056@uni.sydney.edu.au}}
  \href{mailto:your.email@sydney.edu.au}
  {\texttt{cshi0459@uni.sydney.edu.au}}
  \vspace{0.2cm}
}

\checkdata[Date]{
  November 26, 2026
}

\begin{document}
\maketitle
\vspace{-1.1cm}

\section{Introduction}

\begin{figure}
    \centering
    \includegraphics[width=0.98\linewidth]
        {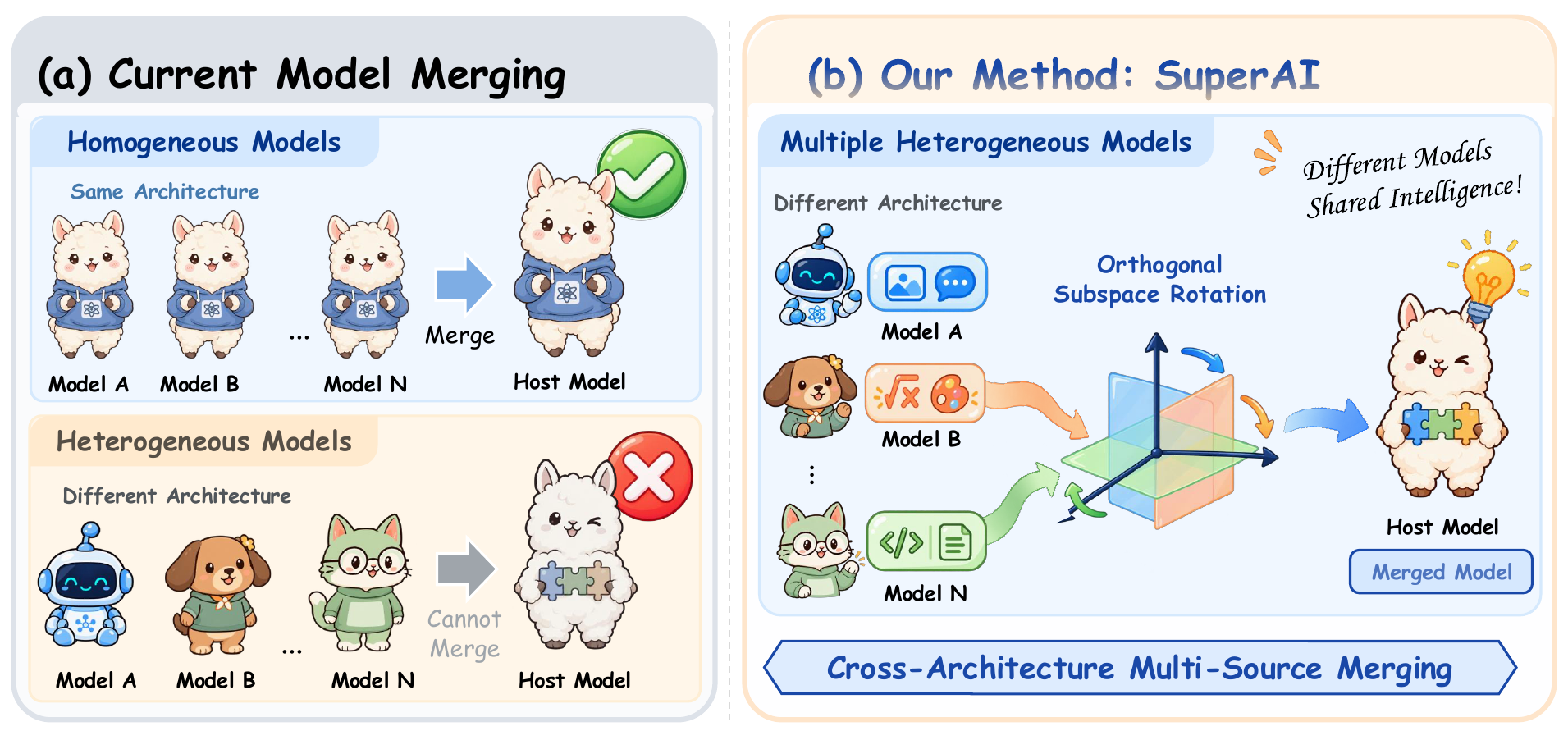}
        \vspace{-0.3cm}
\caption{\textbf{Model Merging \emph{vs} the Proposed SuperAI.} (a) Conventional parameter merging requires compatible architectures. (b) SuperAI integrates heterogeneous source capabilities into one host through orthogonal subspace alignment.}
    \label{fig:ques}
    \vspace{-0.5cm}
\end{figure}

Foundation models~\citep{qwen2.5,olmo20242,team2025gemma} develop complementary capabilities shaped by diverse training data, objectives, and architectures.
These capabilities remain distributed across models~\citep{ramtoula2025fantastic,wei2026optmerge}, while acquiring new ones through further training is costly.
Reusing and composing existing capabilities~\citep{akiba2025evolutionary} therefore offers an important path toward scalable artificial intelligence (AI) systems.

Existing approaches~\citep{wortsman2022model,ilharco2022editing,yoshida2026distac} reuse capabilities through parameter merging~\citep{akiba2025evolutionary}, knowledge distillation~\citep{pmlr-p}, and multi-model collaboration~\citep{wang-etal-2025-mixllm}.
Conventional parameter merging combines weights~\cite{yadav2023ties} or task vectors~\cite{ortiz2023task} but generally requires compatible architectures and parameter structures.
Knowledge distillation transfers information from source-model outputs or intermediate representations through additional host-model training, while routing and collaboration invoke different models according to the input and typically retain multiple models at deployment. To accommodate architectural differences, cross-model fusion methods also use neuron matching~\citep{xu2024training}, optimal transport~\citep{cui2026transport,imfeld2024transformer}, and representation alignment~\citep{dery2026latent} to establish correspondences.
However, as shown in Figure~\ref{fig:ques} (a), integrating complementary information from multiple heterogeneous source models into a common host representation space remains challenging, particularly when improving target-task performance must be balanced with retaining the host model's existing capabilities.

We hypothesize that heterogeneous foundation models can learn structurally corresponding task representations.
Using shared anchor samples, we observe that these representations can be approximately aligned through orthogonal mappings in low-dimensional functional subspaces while preserving each source subspace's geometry.
However, alignment alone does not guarantee task improvements, as transferred information must also be compatible with downstream host computation.

To address these challenges, we propose SuperAI, a cross-architecture multi-model capability fusion framework that integrates complementary capabilities from heterogeneous source models into a single host through orthogonal subspace alignment and lightweight residual injection from Figure~\ref{fig:ques} (b).
Specifically, we first present an orthogonal subspace alignment module, which constructs low-dimensional subspaces of equal dimensionality by applying principal component analysis (PCA) to paired source and host representations of shared anchor samples and aligns their coordinates through orthogonal Procrustes mappings while preserving each source subspace's internal geometry.
Next, we devise a feature mapping module, which defines target increments as the differences between aligned source representations reconstructed in the host hidden space and the original host representations.
Finally, we introduce a multi-source joint training strategy, which keeps the host frozen and jointly supervises a shared lightweight residual multilayer perceptron (MLP) with ground-truth answers and target increments to predict and inject input-dependent residual updates.
The target increments guide residual learning with source-model information, while ground-truth supervision adapts the injected updates to the host's task objectives, connecting representation alignment to task-level capability fusion. Our main contributions are as follows:
\begin{itemize}
    \item We propose SuperAI, a fusion framework combining orthogonal subspace alignment with lightweight residual injection to reuse and compose heterogeneous models' complementary capabilities in a single host.

    \item We identify approximately orthogonal cross-model correspondences in low-dimensional task-relevant subspaces to support capability fusion through host-space target increments and joint task and increment supervision.

    \item Experiments on three host models across 15 capability domains show higher macro-average scores than the evaluated state-of-the-art fusion baselines and better MMLU retention as the number of source models increases.
\end{itemize}

\section{Related Work}
\label{sec:related}

\noindent\textbf{Model Merging.}
Parameter-space merging combines models with compatible architectures through weight averaging or task-vector addition, as in model soups and task arithmetic~\citep{wortsman2022model,ilharco2022editing}.
Subsequent methods mitigate task interference through sign resolution, filtering, conditioning, or loss balancing~\citep{yadav2023ties,qiu2026null,yoshida2026distac}, and improve efficiency through low-rank, orthogonal, or shared parameter structures~\citep{crisostomi2026mass,lee2026adarank,yang2026orthogonal}.
Beyond parameter merging, knowledge fusion transfers information through output distributions or token alignment~\citep{wan2024knowledge,wan2025fusechat}, while neuron and component matching reduces representational mismatch before fusion~\citep{ainsworth2022git,singh2020model,imfeld2024transformer}.
For cross-architecture fusion, Transport and Merge aligns activations and layers to guide weight fusion~\citep{cui2026transport}, while THESEUS transfers task vectors across model widths through functional matching and orthogonal Procrustes alignment~\citep{rinaldi2026transporting}.
However, existing methods struggle to integrate complementary capabilities from multiple heterogeneous sources while preserving the host model's existing capabilities.

\noindent\textbf{Representation Alignment.}
SVCCA and CKA quantify statistical similarity between model representations~\citep{raghu2017svcca,kornblith2019similarity}, while model stitching evaluates their functional compatibility~\citep{lenc2015understanding,bansal2021revisiting,athanasiadis2025grounding}.
Relative representations and latent-space mappings support representation reuse~\citep{moschella2022relative,maiorca2023latent,moayeri2023text,fumero2024latent}, and the Platonic Representation Hypothesis suggests shared structure across independently trained models~\citep{huh2024platonic}.
Recent studies transfer interventions, probes, and adapters across language models~\citep{oozeer2025activation,chen2025transferring} and extend alignment to multiple models through a shared reference space~\citep{achara2026multi}.
However, existing alignment methods establish representational correspondences but provide limited guidance on adapting aligned source information to the host model's downstream computation and task objectives.

\begin{figure*}[t]
\centering
\vspace{-0.3cm}
\includegraphics[width=0.98\linewidth]{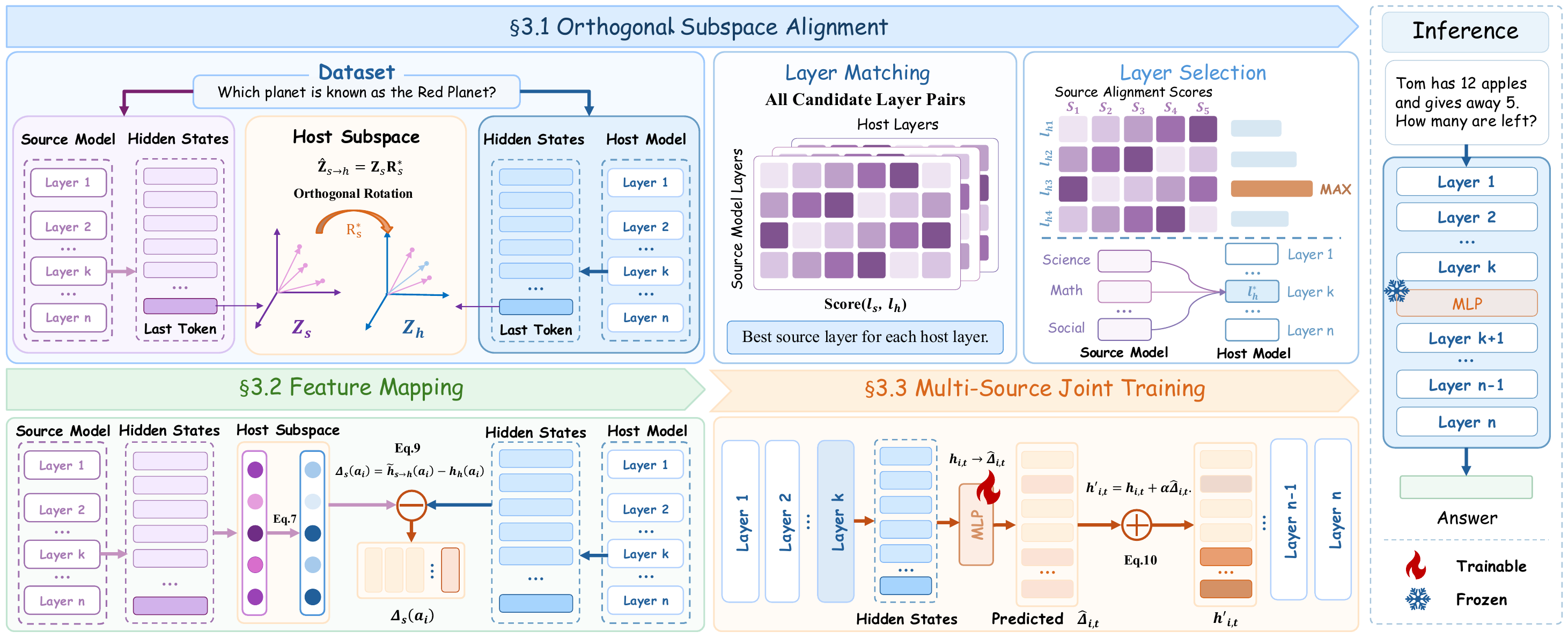}
\vspace{-0.3cm}
\caption{\textbf{Overview of the Proposed SuperAI.}
Shared anchor prompts are used for PCA and orthogonal alignment to select a common host layer and corresponding source layers.
Feature mapping constructs target increments $\Delta_s(a_i)$ in the host hidden space to supervise a MLP jointly with ground-truth answers.
At inference, only the frozen host and trained MLP are required.}
\label{fig:pipeline}
\vspace{-0.6cm}
\end{figure*}

\section{Methodology}

\noindent\textbf{Overview.}
Given a host model $f_h$ and $K$ heterogeneous source models
$\{f_s\}_{s=1}^{K}$, SuperAI learns a MLP to
integrate complementary source capabilities while keeping the
host parameters frozen.
From Figure~\ref{fig:pipeline}, the framework comprises
three components.
First, orthogonal subspace alignment uses shared anchor
prompts to align paired source and host representations in
low-dimensional PCA subspaces of equal dimensionality and selects
a common host layer with source-specific layers and orthogonal
mappings.
Second, feature mapping reconstructs the aligned source
representations in the host hidden space and defines their
differences from the original host representations as target
increments.
Third, multi-source joint training jointly supervises the
MLP with ground-truth answers and target increments to
predict input-dependent residual updates injected at the selected
host layer.
At inference, only the host model and the trained MLP are required.

\subsection{Orthogonal Subspace Alignment}
\label{sec:orthogonal_alignment}

\noindent\textbf{Paired Representations.}
Let $f_h$ be the host model and $\{f_s\}_{s=1}^{K}$ be $K$
heterogeneous source models, with hidden dimensions $d_h$ and $d_s$,
respectively. For each source $s$, shared anchor prompts
$\mathcal{A}_s=\{a_i\}_{i=1}^{n_s}$ provide paired source and host
representations. Let $h_m^{(\ell_m)}(a_i)\in\mathbb{R}^{d_m}$ denote
the hidden state of the last non-padding token at layer $\ell_m$
of model $m\in\{s,h\}$. We stack their transposes as rows of
$H_m^{(\ell_m)}\in\mathbb{R}^{n_s\times d_m}$.
Pairing is defined by the input prompt rather than token position,
allowing the models to use different tokenizers.
For each source--host pair, calibration prompts are used to fit
alignment statistics and mappings, and a separate validation set
$\mathcal{V}_s$ is used for layer selection. Test data are held out
from both fitting and selection.

\noindent\textbf{Subspace Construction.}
For a source model and the host model, we center and normalize the hidden states:
\begin{equation}
u_m(a_i)=
\frac{h_m^{(\ell_m)}(a_i)-\mu_m}
{\max\!\left(
\|h_m^{(\ell_m)}(a_i)-\mu_m\|_2,\epsilon
\right)},
\quad m\in\{s,h\},
\label{eq:superai_normalization}
\end{equation}
where $\mu_m$ is the mean hidden state estimated from the calibration
prompts and $\epsilon>0$ prevents division by zero.
Let $U_m$ stack $u_m(a_i)^\top$ as rows.
We apply PCA separately to the normalized source and host
representations to obtain subspaces of equal dimension $r$:
\begin{equation}
Z_m=(U_m-\mathbf{1}c_m^\top)P_m,
\qquad P_m^\top P_m=I_r,
\label{eq:superai_pca}
\end{equation}
where $c_m\in\mathbb{R}^{d_m}$ is the PCA mean,
$P_m\in\mathbb{R}^{d_m\times r}$ contains the principal components,
and $\mathbf{1}\in\mathbb{R}^{n_s}$ is the all-ones vector.
The coordinate vector
$z_m(a_i)=P_m^\top(u_m(a_i)-c_m)\in\mathbb{R}^{r}$
is the transpose of the corresponding row of $Z_m$.
Both $c_m$ and $P_m$ are fitted on the calibration prompts.
All statistics are source--host-pair specific, including the host-side
statistics; we suppress this dependence and the candidate layer
indices below for clarity.

\noindent\textbf{Orthogonal Alignment.}
Although $Z_s$ and $Z_h$ have the same dimensionality, their
coordinate axes may differ. We align them through orthogonal
Procrustes:
\begin{equation}
\begin{aligned}
R_s^*
&=\underset{R^\top R=I_r}{\arg\min}
\|Z_{s,\mathrm{cal}}R-Z_{h,\mathrm{cal}}\|_F^2,\\
Z_{s,\mathrm{cal}}^\top Z_{h,\mathrm{cal}}
&=L\Sigma V^\top,
\qquad R_s^*=LV^\top,
\end{aligned}
\label{eq:superai_procrustes}
\end{equation}
where $Z_{m,\mathrm{cal}}$ contains calibration coordinates.
The aligned source coordinates are $\widehat{Z}_{s\to h}=Z_sR_s^*$.
Orthogonality preserves pairwise distances within the source PCA subspace.

\noindent\textbf{Layer Selection.}
We evaluate each candidate layer pair on $\mathcal{V}_s$ using
paired cosine similarity and normalized reconstruction error:
\begin{equation}
\begin{aligned}
C_{\mathrm{pair}}
&=\frac{1}{|\mathcal{V}_s|}
\sum_{i\in\mathcal{V}_s}
\cos\!\left((R_s^*)^\top z_s(a_i),z_h(a_i)\right),\\
E_{\mathrm{nrmse}}
&=\frac{
\|Z_{s,\mathcal{V}_s}R_s^*-Z_{h,\mathcal{V}_s}\|_F
}{
\|Z_{h,\mathcal{V}_s}\|_F
}.
\end{aligned}
\label{eq:superai_alignment_metrics}
\end{equation}
Let $C_{\mathrm{CKA}}$ denote linear CKA between
$Z_{s,\mathcal{V}_s}$ and $Z_{h,\mathcal{V}_s}$.
We define the layer-selection score, termed Geometry Match,
as
\begin{equation}
\begin{aligned}
S_s(\ell_s,\ell_h)
=\frac{1}{3}\Big[
&\max(0,C_{\mathrm{pair}})
+\operatorname{clip}(C_{\mathrm{CKA}},0,1)\\
&+\exp(-E_{\mathrm{nrmse}})
\Big].
\end{aligned}
\label{eq:superai_geometry_match}
\end{equation}
These terms measure paired coordinate agreement, structural
similarity, and reconstruction consistency, respectively.
For each candidate host layer, we first select the best source
layer and then aggregate the scores across sources:
\begin{equation}
\begin{aligned}
\ell_s^*(\ell_h)
&=\underset{\ell_s}{\arg\max}\,S_s(\ell_s,\ell_h),\\
\ell_h^*
&=\underset{\ell_h}{\arg\max}
\sum_{s=1}^{K}
S_s\!\left(\ell_s^*(\ell_h),\ell_h\right).
\end{aligned}
\label{eq:superai_layer_selection}
\end{equation}
We write $\ell_s^*=\ell_s^*(\ell_h^*)$ for the final source layer.
Thus, each source has its own selected layer and orthogonal map,
while all sources share the host layer $\ell_h^*$.

\subsection{Feature Mapping}
\label{sec:feature_mapping}

\noindent\textbf{Feature Reconstruction.}
We use the selected source layer $\ell_s^*$, host layer $\ell_h^*$,
and orthogonal map $R_s^*$ to express source information in the
host hidden space.
For a prompt $a_i$, the aligned coordinate vector and its
reconstruction are
\begin{equation}
\begin{aligned}
\widehat{z}_{s\to h}(a_i)
&=(R_s^*)^\top z_s(a_i),\\
b_{s\to h}(a_i)
&=P_h\widehat{z}_{s\to h}(a_i)+c_h.
\end{aligned}
\label{eq:superai_reconstruction}
\end{equation}
Here, $P_h$ and $c_h$ are the host PCA parameters fitted for source
$s$ at the selected host layer.
Because alignment operates on normalized representations, we use
$b_{s\to h}(a_i)$ to specify the direction and the original host
representation to set the centered magnitude:
\begin{equation}
\begin{aligned}
\widetilde{h}_{s\to h}(a_i)
=\mu_h
&+\|h_h(a_i)-\mu_h\|_2\\
&\cdot
\frac{b_{s\to h}(a_i)}
{\max\!\left(\|b_{s\to h}(a_i)\|_2,\epsilon\right)}.
\end{aligned}
\label{eq:superai_mapped_representation}
\end{equation}
Here, $h_h(a_i)=h_h^{(\ell_h^*)}(a_i)$ is the original host
representation at the last non-padding prompt token, and $\mu_h$
is the corresponding calibration mean for source $s$.

\noindent\textbf{Target Increments.}
We define the target increment as the difference between the mapped
source representation and the original host representation:
\begin{equation}
\Delta_s(a_i)
=\widetilde{h}_{s\to h}(a_i)-h_h(a_i)
\in\mathbb{R}^{d_h}.
\label{eq:superai_target_increment}
\end{equation}
This increment expresses the mapped source--host difference in
the host hidden space and provides a representation-level target
for residual learning.
It is used only during training; at inference, the MLP
predicts residual updates directly from host hidden states.

\subsection{Multi-Source Joint Training}
\label{sec:joint_training}

\noindent\textbf{Residual Injection.}
We pool training prompts and ground-truth answers from all source
domains and train one MLP at the selected host
layer $\ell_h^*$.
For each pooled training prompt $a_i$, let $s_i$ identify the source
model used to construct its target increment.
The host parameters remain frozen throughout training.
Given the pre-injection host hidden state
$h_{i,t}\in\mathbb{R}^{d_h}$ at token position $t$, the MLP predicts
a residual update:
\begin{equation}
\begin{aligned}
q_{i,t}
&=\operatorname{SiLU}
\!\left(W_{\mathrm{down}}\operatorname{LN}(h_{i,t})\right),\\
\widehat{\Delta}_{i,t}
&=W_{\mathrm{up}}q_{i,t},\\
h'_{i,t}
&=h_{i,t}+\alpha\widehat{\Delta}_{i,t},
\end{aligned}
\label{eq:superai_residual_injection}
\end{equation}
where $W_{\mathrm{down}}\in\mathbb{R}^{b\times d_h}$,
$W_{\mathrm{up}}\in\mathbb{R}^{d_h\times b}$, and $b$ is the
bottleneck width.
The scaling factor $\alpha$ controls the injected residual strength.
We initialize $W_{\mathrm{up}}$ to zero, yielding zero initial
residual updates.
The MLP operates at all token positions, whereas increment
supervision is applied only at $p_i$, the last non-padding token
position of prompt $a_i$ in the host model.
This matches the position used to construct $\Delta_{s_i}(a_i)$.

\noindent\textbf{Joint Supervision.}
Let $\mathcal{A}^{\mathrm{tr}}_s$ denote the training prompts
associated with source $s$, and let $N$ be the total number of pooled
training prompts.
To account for differences in target magnitude across sources,
we compute the per-source normalization factor
\begin{equation}
\rho_s=
\sqrt{
\frac{1}{|\mathcal{A}^{\mathrm{tr}}_s|d_h}
\sum_{a_i\in\mathcal{A}^{\mathrm{tr}}_s}
\|\Delta_s(a_i)\|_2^2
}.
\label{eq:superai_increment_scale}
\end{equation}
The increment loss supervises the unscaled MLP prediction:
\begin{equation}
\mathcal{L}_{\mathrm{inc}}
=
\frac{1}{Nd_h}
\sum_{i=1}^{N}
\left\|
\frac{
\widehat{\Delta}_{i,p_i}-\Delta_{s_i}(a_i)
}{
\max(\rho_{s_i},\epsilon)
}
\right\|_2^2.
\label{eq:superai_increment_loss}
\end{equation}
Thus, $\Delta_{s_i}(a_i)$ supervises
$\widehat{\Delta}_{i,p_i}$, while $\alpha$ separately controls the
strength of the update applied to the host.
We additionally supervise task outputs using ground-truth answers.
Let $y_i$ be the answer to prompt $a_i$ and $\mathcal{T}_i$ its
answer-token positions.
The task loss is
\begin{equation}
\mathcal{L}_{\mathrm{task}}
=
-\frac{1}{N}\sum_{i=1}^{N}
\frac{1}{|\mathcal{T}_i|}
\sum_{t\in\mathcal{T}_i}
\log p_{\theta}(y_{i,t}\mid a_i,y_{i,<t}),
\label{eq:superai_task_loss}
\end{equation}
where $p_{\theta}$ is the output distribution of the host augmented
with the MLP, and $\theta$ denotes the trainable MLP
parameters.
The joint objective is
\begin{equation}
\mathcal{L}
=
\lambda_{\mathrm{task}}\mathcal{L}_{\mathrm{task}}
+
\lambda_{\mathrm{inc}}\mathcal{L}_{\mathrm{inc}},
\label{eq:superai_joint_loss}
\end{equation}
where $\lambda_{\mathrm{task}}$ and $\lambda_{\mathrm{inc}}$
weight the two supervision signals.
Increment supervision guides residual learning with mapped source
information, while task supervision adapts the injected updates
to the host's task objectives.

\noindent\textbf{Inference Procedure.}
At inference, the MLP predicts input-dependent residual
updates from the host hidden states and injects them at
$\ell_h^*$ using Eq.~\ref{eq:superai_residual_injection}.
Only the host model and the trained MLP are retained.
The source identity $s_i$ is used only to select training targets
and normalization factors; source models, source identities,
and explicit feature mappings are not required at inference.

\section{Experiments}

\begin{table*}[t]
\centering
\vspace{-0.3cm}
\caption{\textbf{Quantitative Comparison of Multi-Source Capability Fusion across Different Host Models.}
We report the original host performance and multi-source fusion results across 15 capability domains. The best and second-best results among the fusion methods are shown in \textbf{bold} and \underline{underlined}, respectively. Green and red arrows in the Avg. column denote gains and decreases relative to the corresponding original host. Column abbreviations and complete source--benchmark--metric mappings are provided in the appendix.}
\label{tab:multi_host_capability_merging}
\vspace{-0.2cm}

\scriptsize

\resizebox{\linewidth}{!}{
\begin{tabular}{l*{15}{c}@{\hspace{4pt}}c}
\toprule
\textbf{Method}
& \textbf{Math}
& \textbf{Safety}
& \textbf{Law}
& \textbf{Phys.}
& \textbf{Sci.}
& \textbf{C.Sense}
& \textbf{Social}
& \textbf{Code}
& \textbf{Multi.}
& \textbf{Fin./Bus.}
& \textbf{Astro.}
& \textbf{Med.}
& \textbf{Climate}
& \textbf{Cyber}
& \textbf{Chem.}
& \textbf{Avg.} \\
\midrule

\textcolor{gray}{\textit{Domain}}
& \textcolor{gray}{\textit{67.7}}
& \textcolor{gray}{\textit{72.0}}
& \textcolor{gray}{\textit{46.5}}
& \textcolor{gray}{\textit{80.9}}
& \textcolor{gray}{\textit{76.5}}
& \textcolor{gray}{\textit{81.6}}
& \textcolor{gray}{\textit{75.2}}
& \textcolor{gray}{\textit{81.5}}
& \textcolor{gray}{\textit{69.4}}
& \textcolor{gray}{\textit{15.0}}
& \textcolor{gray}{\textit{74.3}}
& \textcolor{gray}{\textit{58.4}}
& \textcolor{gray}{\textit{71.4}}
& \textcolor{gray}{\textit{36.2}}
& \textcolor{gray}{\textit{49.0}}
& \textcolor{gray}{\textit{63.7}} \\

\midrule
\multicolumn{17}{l}{%
    \cellcolor{gray!7}\textit{Host model: Llama-3.2-3B}} \\

\textcolor{gray}{\textit{Original}}
& \textcolor{gray}{16.7}
& \textcolor{gray}{56.8}
& \textcolor{gray}{38.0}
& \textcolor{gray}{68.3}
& \textcolor{gray}{67.9}
& \textcolor{gray}{66.2}
& \textcolor{gray}{58.7}
& \textcolor{gray}{39.7}
& \textcolor{gray}{56.1}
& \textcolor{gray}{3.7}
& \textcolor{gray}{60.5}
& \textcolor{gray}{50.9}
& \textcolor{gray}{42.3}
& \textcolor{gray}{31.4}
& \textcolor{gray}{44.0}
& \textcolor{gray}{46.7} \\

THESEUS~\citep{rinaldi2026transporting}
& \underline{24.1}
& 57.3
& 40.3
& \underline{62.9}
& 63.5
& 60.0
& 53.8
& \textbf{42.1}
& 44.0
& \underline{12.7}
& \underline{54.2}
& \underline{40.6}
& \underline{38.4}
& \underline{34.7}
& 32.1
& \underline{44.0}
\ \textcolor{red!65!black}{($\downarrow 2.7$)} \\

T\&M~\citep{cui2026transport}
& 17.6
& \underline{59.5}
& \textbf{41.2}
& 61.4
& \underline{64.8}
& \underline{61.7}
& \underline{54.2}
& \underline{41.7}
& \underline{44.3}
& 10.9
& 49.7
& 40.3
& 35.8
& 31.2
& \underline{43.4}
& 43.8
\ \textcolor{red!65!black}{($\downarrow 2.9$)} \\

\rowcolor{blue!6}
\textbf{SuperAI}
& \textbf{40.7}
& \textbf{70.9}
& \underline{40.8}
& \textbf{89.1}
& \textbf{69.8}
& \textbf{68.2}
& \textbf{63.3}
& 28.3
& \textbf{55.0}
& \textbf{14.7}
& \textbf{57.9}
& \textbf{48.2}
& \textbf{72.5}
& \textbf{48.9}
& \textbf{44.0}
& \textbf{54.2}
\ \textcolor{green!45!black}{($\uparrow 7.5$)} \\

\midrule
\multicolumn{17}{l}{%
    \cellcolor{gray!7}\textit{Host model: Qwen2.5-3B}} \\

\textcolor{gray}{\textit{Original}}
& \textcolor{gray}{26.3}
& \textcolor{gray}{61.3}
& \textcolor{gray}{39.4}
& \textcolor{gray}{85.1}
& \textcolor{gray}{73.0}
& \textcolor{gray}{77.2}
& \textcolor{gray}{69.0}
& \textcolor{gray}{55.6}
& \textcolor{gray}{71.9}
& \textcolor{gray}{7.0}
& \textcolor{gray}{71.2}
& \textcolor{gray}{53.6}
& \textcolor{gray}{68.1}
& \textcolor{gray}{39.2}
& \textcolor{gray}{45.0}
& \textcolor{gray}{56.2} \\

THESEUS~\citep{rinaldi2026transporting}
& 22.7
& 55.6
& 40.5
& \underline{71.8}
& 56.2
& 54.0
& 59.1
& 41.5
& 57.8
& \textbf{13.4}
& 46.7
& 37.2
& 31.9
& 32.6
& 29.0
& 43.3
\ \textcolor{red!65!black}{($\downarrow 12.9$)} \\

T\&M~\citep{cui2026transport}
& \underline{31.2}
& \underline{61.6}
& \underline{43.8}
& 65.9
& \underline{63.7}
& \underline{62.8}
& \underline{65.9}
& \textbf{53.8}
& \underline{64.9}
& \underline{12.7}
& \underline{55.0}
& \underline{43.4}
& \underline{40.8}
& \underline{34.1}
& \underline{35.6}
& \underline{49.0}
\ \textcolor{red!65!black}{($\downarrow 7.2$)} \\

\rowcolor{blue!6}
\textbf{SuperAI}
& \textbf{45.7}
& \textbf{76.6}
& \textbf{53.5}
& \textbf{94.9}
& \textbf{82.5}
& \textbf{79.5}
& \textbf{69.5}
& \underline{44.2}
& \textbf{67.6}
& 9.3
& \textbf{72.4}
& \textbf{53.8}
& \textbf{73.8}
& \textbf{53.8}
& \textbf{43.0}
& \textbf{61.3}
\ \textcolor{green!45!black}{($\uparrow 5.1$)} \\

\midrule
\multicolumn{17}{l}{%
    \cellcolor{gray!7}\textit{Host model: Pythia-2.8B}} \\

\textcolor{gray}{\textit{Original}}
& \textcolor{gray}{3.7}
& \textcolor{gray}{42.5}
& \textcolor{gray}{23.9}
& \textcolor{gray}{24.2}
& \textcolor{gray}{24.0}
& \textcolor{gray}{20.7}
& \textcolor{gray}{35.7}
& \textcolor{gray}{11.1}
& \textcolor{gray}{23.6}
& \textcolor{gray}{0.7}
& \textcolor{gray}{28.3}
& \textcolor{gray}{26.2}
& \textcolor{gray}{27.9}
& \textcolor{gray}{23.3}
& \textcolor{gray}{27.0}
& \textcolor{gray}{22.9} \\

THESEUS~\citep{rinaldi2026transporting}
& 1.8
& 13.4
& 4.3
& \underline{54.9}
& 20.1
& \underline{31.9}
& 25.5
& \underline{3.8}
& 6.2
& \underline{6.9}
& 18.2
& 7.9
& 38.9
& \underline{32.7}
& 20.5
& 19.1
\ \textcolor{red!65!black}{($\downarrow 3.8$)} \\

T\&M~\citep{cui2026transport}
& \underline{3.1}
& \underline{20.1}
& \underline{6.0}
& 47.5
& \underline{27.1}
& \textbf{40.6}
& \underline{32.2}
& \textbf{6.0}
& \underline{9.4}
& \textbf{10.3}
& \underline{21.8}
& \underline{11.6}
& \underline{46.0}
& 29.2
& \textbf{23.1}
& \underline{22.3}
\ \textcolor{red!65!black}{($\downarrow 0.6$)} \\

\rowcolor{blue!6}
\textbf{SuperAI}
& \textbf{8.0}
& \textbf{69.9}
& \textbf{31.0}
& \textbf{89.2}
& \textbf{27.6}
& 18.7
& \textbf{37.5}
& 1.3
& \textbf{29.3}
& 1.0
& \textbf{25.0}
& \textbf{27.7}
& \textbf{72.5}
& \textbf{33.0}
& \underline{21.0}
& \textbf{32.8}
\ \textcolor{green!45!black}{($\uparrow 9.9$)} \\

\bottomrule
\end{tabular}}
\vspace{-0.3cm}
\end{table*}

\noindent\textbf{Source Models and Datasets. }
We select three model families as host models, covering multiple parameter scales within each family: the Llama family includes Llama-3.2-1B, Llama-3.2-3B, and Llama-3.1-8B~\cite{grattafiori2024llama}; the Pythia family includes Pythia-1B, Pythia-2.8B, Pythia-6.9B, and Pythia-12B~\cite{biderman2023pythia}; and the Qwen family includes Qwen2.5-1.5B, Qwen2.5-3B, Qwen2.5-7B, and Qwen2.5-14B~\cite{qwen2.5}. We construct a pool of 15 source models spanning science, commonsense and social reasoning, math, code generation, multilingual reasoning, safety, law, finance, astronomy, medicine, climate science, cybersecurity, chemistry, and physics. The pool includes four OFT sources~\cite{yang2026orthogonal,qiu2023controlling}, Qwen2.5-Coder-7B-Instruct~\cite{hui2024qwen2}, two MergeBench source models~\cite{he2026mergebench}, and eight domain-specialized source models. These source models vary in model family, parameter scale, and training procedure, forming a heterogeneous testbed for cross-architecture capability transfer and multi-source fusion. More details are in the appendix.

\noindent\textbf{Evaluation Metrics. }
We use one predefined primary metric for each domain. Most domains are evaluated with accuracy or task-specific variants of accuracy. The main exceptions are code generation and cybersecurity: MBPP~\cite{austin2021program} is evaluated with Pass@1~\cite{chen2021evaluating}, requiring the generated program to pass the test suite, while SEVENLLM~\cite{ji2024sevenllm} uses token-level F1. All metrics are higher-is-better. For multi-source fusion, we additionally report the unweighted macro average across 15 domains and MMLU~\cite{hendrycks2020measuring} to assess general-capability retention as more source models are merged. For representation-level analysis, we additionally report the mapped--host cosine score $C_{\mathrm{map}}$, which measures the average cosine similarity between the mapped source representation and the host representation after injecting the MLP-predicted residual.

\noindent\textbf{Hyperparameter.}
We use Llama-3.2-3B~\cite{grattafiori2024llama}, Qwen2.5-3B~\cite{qwen2.5}, and Pythia-2.8B~\cite{biderman2023pythia} as host models. For each source-host pair, we extract hidden representations from shared prompts and perform PCA followed by orthogonal Procrustes alignment to establish cross-model correspondences. Both the PCA subspace dimension and the bottleneck dimension of the MLP are set to 128. During training, the host model remains frozen, and only the MLP is optimized. We train for two epochs using AdamW with a batch size of 4, a learning rate of $2 \times 10^{-4}$, and a weight decay of 0.01. For each configuration, we jointly train on pooled data from the selected domains, starting independently from the original host model. THESEUS~\cite{rinaldi2026transporting} uses activation-based calibration and validation-based scaling, while Transport and Merge (T\&M)~\cite{cui2026transport} applies post-fusion adaptation, following their original protocols. Detailed dataset configurations, loss weights, and the residual scaling factor are provided in the appendix.

\subsection{Main Results}

\noindent\textbf{Multi-Source Fusion.}
To evaluate SuperAI across host architectures, we jointly fuse all 15 domain source models into Llama-3.2-3B, Qwen2.5-3B, and Pythia-2.8B and compare the models with THESEUS~\cite{rinaldi2026transporting} and Transport and Merge~\cite{cui2026transport}.
Table~\ref{tab:multi_host_capability_merging} shows that SuperAI achieves the highest macro-average score on all three hosts, raising the scores of Llama, Qwen, and Pythia from 46.7, 56.2, and 22.9 to 54.2, 61.3, and 32.8, respectively.
These correspond to gains of 7.5, 5.1, and 9.9 points over the original hosts and 10.2, 12.3, and 10.5 points over the strongest merging baselines.
Both baselines score below the original hosts on average, suggesting potential capability interference during heterogeneous multi-source fusion.
At the domain level, Llama improves in climate and math, Qwen in math, safety, and law, and Pythia in physics and climate.
Code generation declines across all three hosts, indicating that gains are not uniformly distributed across domains.
Although Pythia records the largest average gain, Qwen achieves the highest final score, showing that larger gains do not necessarily imply stronger post-fusion performance.
These distinct domain-level gain patterns from the same source models suggest that capability absorption depends on source complementarity, the host's initial capabilities, and domain characteristics.
Overall, these results support the effectiveness of SuperAI's orthogonal subspace alignment and residual injection for heterogeneous multi-source fusion and suggest reduced performance interference.

\begin{figure}
    \centering
    \includegraphics[width=0.98\linewidth]{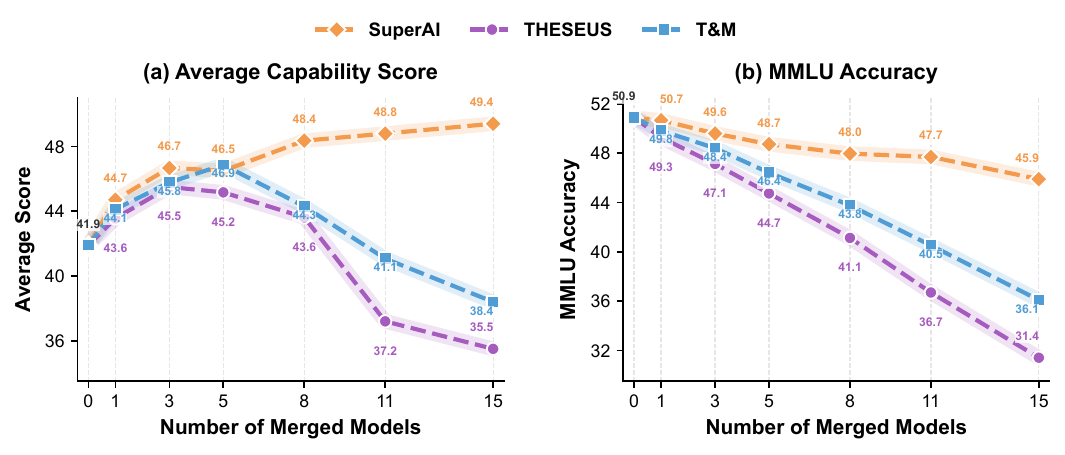}
    \vspace{-0.3cm}
\caption{\textbf{Effect of Source Model Count.}
(a) Average capability scores and (b) MMLU scores as more source models are merged.
Results for SuperAI, THESEUS~\citep{rinaldi2026transporting}, and T\&M~\citep{cui2026transport} are averaged across three host models.}
    \vspace{-0.4cm}
    \label{fig:mmlu}
\end{figure}

\noindent\textbf{General Capability Retention.}
To evaluate capability acquisition and general capability retention during multi-source fusion, we gradually increase the number of integrated source models from 1 to 15 and compare SuperAI, THESEUS~\citep{rinaldi2026transporting}, and Transport and Merge (T\&M)~\citep{cui2026transport} in terms of average scores across 15 domains and MMLU scores.
The value 0 represents the original host model without fusion.
As shown in Figure~\ref{fig:mmlu}(a), the average score of SuperAI generally increases from 41.9 to 49.4 as more source models are integrated.
In contrast, the average scores of THESEUS and T\&M initially increase but then decline, eventually reaching 35.5 and 38.4, respectively.
This suggests that interference becomes more pronounced for the two baselines as additional source models are integrated.
As shown in Figure~\ref{fig:mmlu}(b), MMLU scores decrease for all three methods as the number of source models increases.
After integrating all 15 source models, the MMLU score of SuperAI decreases from 50.9 to 45.9, a drop of 5.0 points, whereas THESEUS and T\&M decrease by 19.5 and 14.8 points, respectively.
These results suggest that SuperAI is more robust to interference than the two baselines, enabling complementary capability acquisition while better preserving the host model's existing general capabilities.

\begin{figure}
    \centering
    \includegraphics[width=0.98\linewidth]{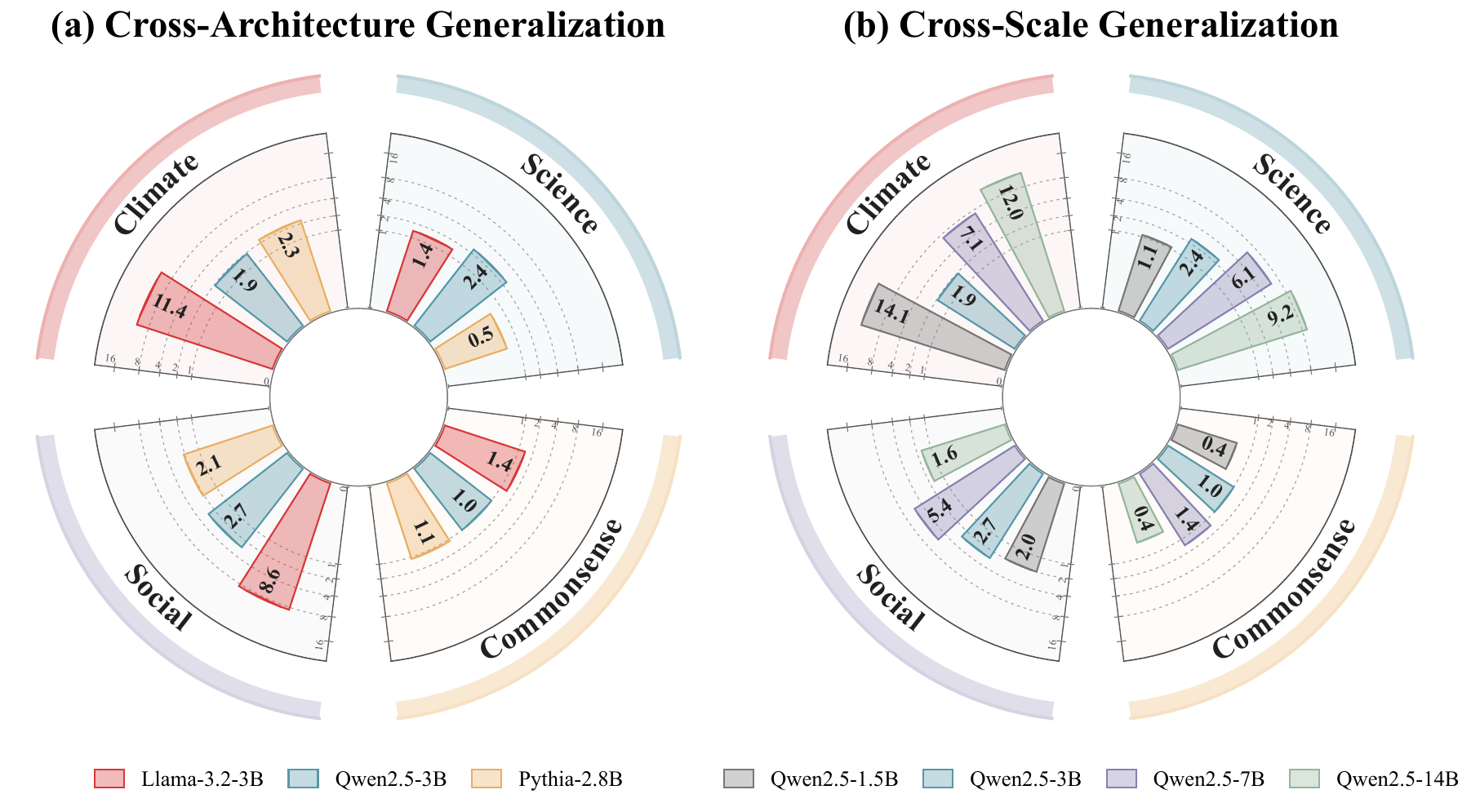}
    \vspace{-0.15cm}
\caption{\textbf{Generalization Across Host Architectures and Scales.}
Performance gains over the original hosts on unseen datasets in Climate, Science, Commonsense, and Social: (a) different host architectures; (b) Qwen2.5 model scales.}
    \label{fig:generalization}
    \vspace{-1em}
\end{figure}

\noindent\textbf{Cross-Dataset Generalization.}
To evaluate SuperAI's generalization, we conduct experiments along three dimensions: cross-architecture, cross-dataset, and cross-scale generalization.
The results in Figure~\ref{fig:generalization}(a) show that Llama-3.2-3B, Qwen2.5-3B, and Pythia-2.8B achieve average gains of 5.7, 2.0, and 1.5 points, respectively, on unseen datasets across Climate, Science, Commonsense, and Social.
Figure~\ref{fig:generalization}(b) shows that the Qwen2.5-1.5B, 3B, 7B, and 14B models achieve average gains of 4.4, 2.0, 5.0, and 5.8 points across the four domains, respectively, with the 7B and 14B models exhibiting stronger overall capability absorption.
In the Science domain, the gain increases from 1.1 to 9.2 points as the model scale grows, indicating that larger host models can more effectively absorb source capabilities in certain domains.
Therefore, SuperAI enables consistent positive transfer across model architectures, unseen data distributions, and parameter scales, supporting orthogonal subspace alignment as a generalizable interface for cross-model capability fusion.
However, domain-specific returns remain non-linear and depend on the host model's existing capabilities and headroom for improvement.

\subsection{Ablation Study}

\begin{table}[t]
\centering
\caption{\textbf{Effect of Orthogonal Subspace Alignment.}
We report 15-domain macro-average scores for three host models, with changes relative to each original host in parentheses. Best results are \textbf{bolded}.}
\vspace{-0.3cm}
\label{tab:alignment_ablation}
\tiny
\setlength{\tabcolsep}{2.4pt}
\renewcommand{\arraystretch}{1.10}
\resizebox{\linewidth}{!}{
\begin{tabular}{lccc}
\toprule
\textbf{Mapping Method}
& \textbf{Llama-3.2-3B}
& \textbf{Qwen2.5-3B}
& \textbf{Pythia-2.8B} \\
\midrule

\textcolor{gray}{Original}
& \textcolor{gray}{46.7}
& \textcolor{gray}{56.2}
& \textcolor{gray}{22.9} \\

No Alignment
& 47.6 \textcolor{green!45!black}{($\uparrow 0.9$)}
& 56.1 \textcolor{red!65!black}{($\downarrow 0.1$)}
& 21.0 \textcolor{red!65!black}{($\downarrow 1.9$)} \\

PCA Only
& 47.8 \textcolor{green!45!black}{($\uparrow 1.1$)}
& 56.9 \textcolor{green!45!black}{($\uparrow 0.7$)}
& 28.6 \textcolor{green!45!black}{($\uparrow 5.7$)} \\

PCA + Linear Alignment
& 53.2 \textcolor{green!45!black}{($\uparrow 6.5$)}
& 57.5 \textcolor{green!45!black}{($\uparrow 1.3$)}
& 27.2 \textcolor{green!45!black}{($\uparrow 4.3$)} \\

\rowcolor{blue!6}
PCA + Orthogonal(Ours)
& \textbf{54.2} \textcolor{green!45!black}{($\uparrow 7.5$)}
& \textbf{61.3} \textcolor{green!45!black}{($\uparrow 5.1$)}
& \textbf{32.8} \textcolor{green!45!black}{($\uparrow 9.9$)} \\

\bottomrule
\end{tabular}
}
\vspace{-1em}
\end{table}

\noindent\textbf{Orthogonal Alignment.} To evaluate the role of orthogonal subspace alignment, we compare No Alignment, PCA Only, PCA with Linear Mapping, and PCA with Orthogonal Mapping under identical training settings. The results in Table~\ref{tab:alignment_ablation} show that PCA with Orthogonal Mapping consistently achieves the best performance across all three host models. Specifically, it improves Llama, Qwen, and Pythia by 7.5, 5.1, and 9.9 points, respectively, whereas the other configurations produce smaller or less consistent gains. It outperforms linear mapping by 1.0, 3.8, and 5.6 points on the three hosts, respectively. Therefore, orthogonal alignment establishes a reliable cross-model interface by reconciling coordinate differences while preserving the geometry of PCA-projected source representations.

\begin{table}[t]
\centering
\caption{\textbf{Effect of Mapper Architecture.}
We report average scores across 15 domains for three host models.
Parentheses indicate score changes relative to the original host.}
\label{tab:mapper_architecture}

\vspace{-0.3cm}
\scriptsize
\setlength{\tabcolsep}{8pt}
\renewcommand{\arraystretch}{1.15}
\resizebox{0.98\linewidth}{!}{%
\begin{tabular}{lccc}
\toprule
\textbf{Mapper}
& \textbf{Llama-3.2-3B}
& \textbf{Qwen2.5-3B}
& \textbf{Pythia-2.8B} \\
\midrule

\textcolor{gray}{Original}
& \textcolor{gray}{46.7}
& \textcolor{gray}{56.2}
& \textcolor{gray}{22.9} \\

Linear Mapper
& 50.6 \textcolor{green!45!black}{($\uparrow 3.9$)}
& 56.0 \textcolor{red!65!black}{($\downarrow 0.2$)}
& 30.1 \textcolor{green!45!black}{($\uparrow 7.2$)} \\

\rowcolor{blue!6}
MLP Mapper(Ours)
& \textbf{54.2} \textcolor{green!45!black}{($\uparrow 7.5$)}
& \textbf{61.3} \textcolor{green!45!black}{($\uparrow 5.1$)}
& \textbf{32.8} \textcolor{green!45!black}{($\uparrow 9.9$)} \\

\bottomrule
\end{tabular}
}
\vspace{-6pt}
\end{table}

\noindent\textbf{Residual Mapping.}
To evaluate whether nonlinear residual mapping is necessary for adapting aligned source representations to host computation, we compare a linear mapper with SuperAI's MLP across three hosts.
Results in Table~\ref{tab:mapper_architecture} show that the linear mapper produces inconsistent performance changes, whereas the MLP consistently achieves larger gains across three models.
The linear mapper improves Llama and Pythia by 3.9 and 7.2 points, respectively, but reduces Qwen performance by 0.2 points.
In contrast, the MLP improves Llama, Qwen, and Pythia by 7.5, 5.1, and 9.9 points, outperforming the linear mapper by 3.6, 5.3, and 2.7 points, respectively.
Representation alignment establishes geometric correspondences across models, but residual mapping is needed to adapt source-model information to the host's downstream computation.
Therefore, the effectiveness of linear mapping for capability transfer varies across host architectures, whereas nonlinear residual mapping learns host-compatible increments and enables consistent capability transfer across architectures.

\begin{table}[t]
\centering
\caption{\textbf{Effect of Training Objectives.}
Mapped-host cosine similarity scores and 15-domain macro-average scores for three host models.
Parentheses show score changes from the original host.}
\label{tab:objective_ablation}
\vspace{-0.2cm}

\scriptsize
\setlength{\tabcolsep}{9pt}
\renewcommand{\arraystretch}{1.10}

\resizebox{0.98\linewidth}{!}{%
\begin{tabular}{lccc}
\toprule

\textbf{Objective}
& \textbf{Llama-3.2-3B}
& \textbf{Qwen2.5-3B}
& \textbf{Pythia-2.8B} \\

\midrule

\rowcolor{gray!7}
\multicolumn{4}{l}{\textit{Mapped-Host Cosine Score}} \\[1pt]

\rowcolor{white}
$\lambda_{\mathrm{inc}}=0$
& 0.341
& 0.443
& 0.692 \\

$\lambda_{\mathrm{task}}=0$
& \underline{0.937}
& \underline{0.925}
& \underline{0.971} \\

\rowcolor{blue!6}
Joint (Ours)
& \textbf{0.947}
& \textbf{0.939}
& \textbf{0.972} \\

\midrule

\rowcolor{gray!7}
\multicolumn{4}{l}{\textit{15-Domain Macro-Average Score}} \\[1pt]

\rowcolor{white}
\textcolor{gray}{Original}
& \textcolor{gray}{46.7}
& \textcolor{gray}{56.2}
& \textcolor{gray}{22.9} \\

$\lambda_{\mathrm{inc}}=0$
& \underline{50.6}\,
  \textcolor{green!45!black}{($\uparrow 3.9$)}
& \underline{57.5}\,
  \textcolor{green!45!black}{($\uparrow 1.3$)}
& \underline{27.1}\,
  \textcolor{green!45!black}{($\uparrow 4.2$)} \\

$\lambda_{\mathrm{task}}=0$
& 44.8\,
  \textcolor{red!65!black}{($\downarrow 1.9$)}
& 56.3\,
  \textcolor{green!45!black}{($\uparrow 0.1$)}
& 24.0\,
  \textcolor{green!45!black}{($\uparrow 1.1$)} \\

\rowcolor{blue!6}
Joint (Ours)
& \textbf{54.2}\,
  \textcolor{green!45!black}{($\uparrow 7.5$)}
& \textbf{61.3}\,
  \textcolor{green!45!black}{($\uparrow 5.1$)}
& \textbf{32.8}\,
  \textcolor{green!45!black}{($\uparrow 9.9$)} \\

\bottomrule
\end{tabular}%
}
\end{table}

\noindent\textbf{Joint Supervision.} To evaluate the roles of task-output and source-model increment supervision, we compare $\lambda_{\mathrm{inc}}=0$, $\lambda_{\mathrm{task}}=0$, and the joint objective across three host models, using the mapped--host cosine score to measure the similarity between the mapped source representation and the host representation after residual injection. The results in Table~\ref{tab:objective_ablation} show that joint supervision achieves the best downstream performance, while maintaining high representation similarity. Specifically, $\lambda_{\mathrm{inc}}=0$ improves Llama, Qwen, and Pythia by 3.9, 1.3, and 4.2 points but yields substantially lower cosine scores. Conversely, $\lambda_{\mathrm{task}}=0$ achieves high mapped--host cosine scores but provides inconsistent downstream improvements. In contrast, the joint objective improves the three hosts by 7.5, 5.1, and 9.9 points. Therefore, joint supervision balances source-model guidance with host task adaptation, improving cross-architecture capability transfer.

\subsection{Deeper Analysis}

\begin{figure}
    \centering
    \includegraphics[width=0.98\linewidth]{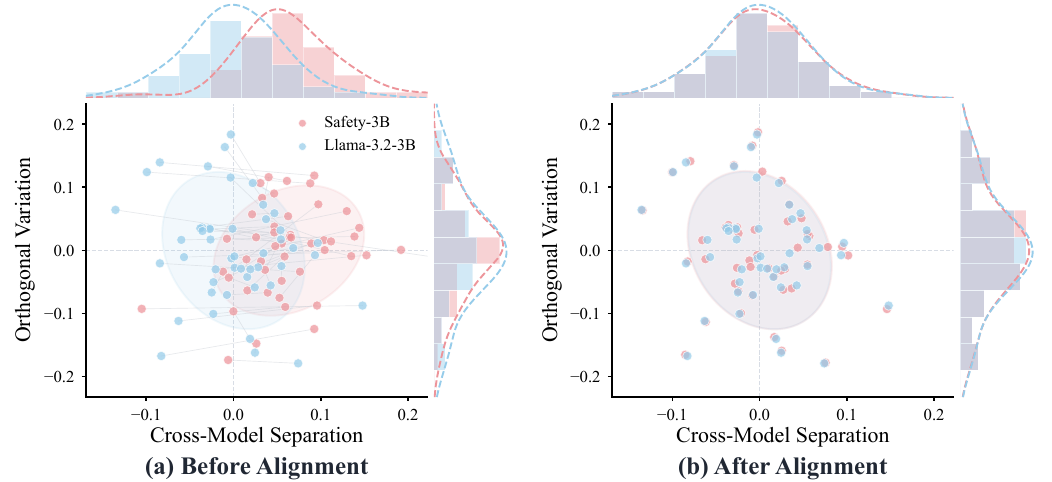}
    \vspace{-0.2cm}
\caption{\textbf{Cross-Model Representation Alignment.}
Source and host distributions overlap more closely after alignment.
The horizontal axis follows their pre-alignment centroid difference, while the vertical axis captures the dominant orthogonal variation.}

    \label{fig:alignment}
    \vspace{-0.3cm}
\end{figure}

\begin{figure}
    \centering
    \includegraphics[width=0.98\linewidth]{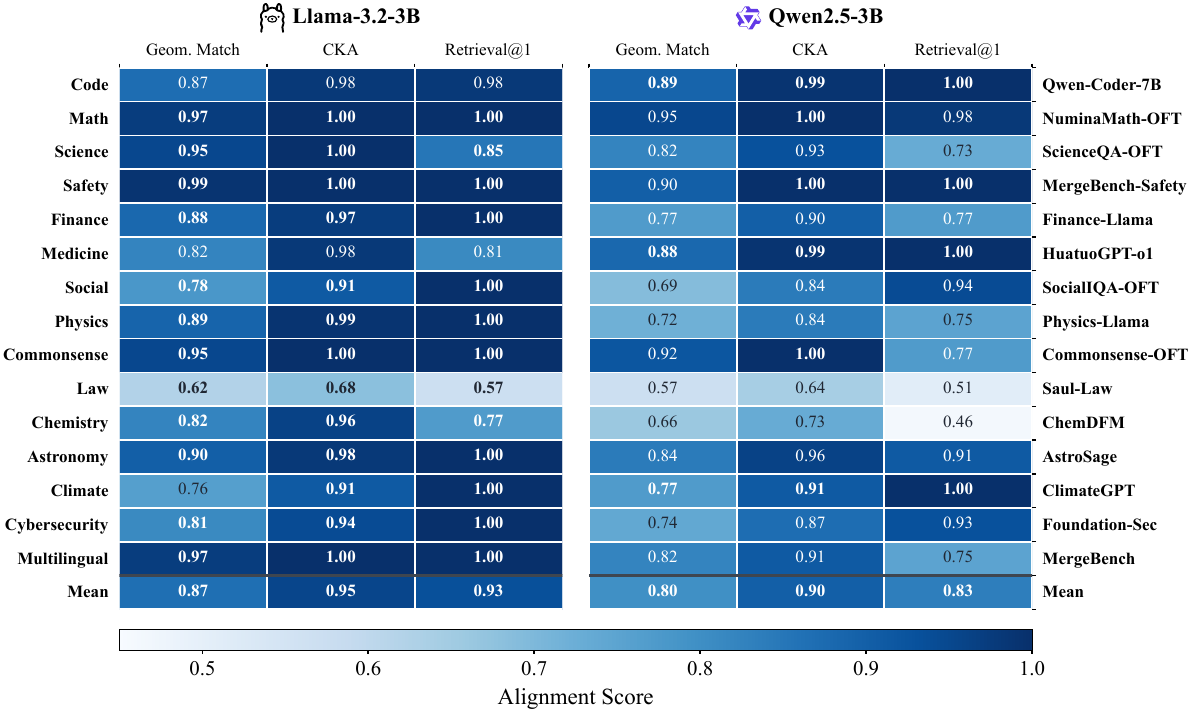}
    \vspace{-0.3cm}
    \caption{\textbf{Cross-Model Alignment Evaluation.} Geometry Match, CKA, and Retrieval@1 scores across 15 source models with Llama and Qwen as host models.}
    \label{fig:merge_compare}
    \vspace{-0.3cm}
\end{figure}

\noindent\textbf{Geometric Correspondence.}
To investigate whether heterogeneous models share alignable task-relevant representations, we first visualize the source and host representations before and after orthogonal alignment.
As shown in Figure~\ref{fig:alignment}(a), the two sets of representations differ substantially in position, orientation, and distribution before alignment.
After applying the learned orthogonal mapping, their point clouds and distribution contours become closely aligned, as shown in Figure~\ref{fig:alignment}(b).
This suggests that heterogeneous models may encode similar task-relevant information in different coordinate systems while preserving underlying geometric structures that can be aligned.
We further evaluate 15 heterogeneous source models with Llama-3.2-3B and Qwen2.5-3B as host models using Geometry Match, linear CKA, and Retrieval@1.
As shown in Figure~\ref{fig:merge_compare}, the average scores are 0.87, 0.95, and 0.93 for Llama-3.2-3B, and 0.80, 0.90, and 0.83 for Qwen2.5-3B, respectively.
The high CKA scores indicate that their representations share similar geometric relationships across samples.
Geometry Match measures the overall compatibility of the selected layer pairs, while Retrieval@1 shows that the learned mappings establish correspondences between individual samples.
The differences between the two host models also suggest that alignment quality depends on the specific combination of source and host models.
Together, these results suggest that task-relevant representations across heterogeneous models have geometric structures that can be aligned.
This provides a representation-level basis for mapping source-model information into the host space and constructing target feature increments.

\begin{table}
\centering
\caption{\textbf{Effect of Layer-Pair Alignment Quality.}
We report 15-domain averages and mapped-host cosine match scores; parentheses show changes from the baseline.}
\label{tab:layer_pair_quality}
\vspace{-0.3cm}

\scriptsize
\setlength{\tabcolsep}{5pt}
\renewcommand{\arraystretch}{1.10}

\resizebox{0.95\linewidth}{!}{%
\begin{tabular}{lccc}
\toprule

\textbf{Layer Pair}
& \textbf{Llama-3.2-3B}
& \textbf{Qwen2.5-3B}
& \textbf{Pythia-2.8B} \\

\midrule

\rowcolor{gray!7}
\multicolumn{4}{l}{\textit{15-Domain Average Score}} \\[1pt]

\rowcolor{white}
\textcolor{gray}{Original}
& \textcolor{gray}{46.7}
& \textcolor{gray}{56.2}
& \textcolor{gray}{22.9} \\

Low Matched
& 51.1\,\textcolor{green!45!black}{($\uparrow 4.4$)}
& 59.0\,\textcolor{green!45!black}{($\uparrow 2.8$)}
& 28.4\,\textcolor{green!45!black}{($\uparrow 5.5$)} \\

Medium Matched
& \underline{53.5}\,\textcolor{green!45!black}{($\uparrow 6.8$)}
& \underline{60.4}\,\textcolor{green!45!black}{($\uparrow 4.2$)}
& \underline{31.6}\,\textcolor{green!45!black}{($\uparrow 8.7$)} \\

\rowcolor{blue!6}
\textbf{High Matched}
& \textbf{54.2}\,\textcolor{green!45!black}{($\uparrow 7.5$)}
& \textbf{61.3}\,\textcolor{green!45!black}{($\uparrow 5.1$)}
& \textbf{32.8}\,\textcolor{green!45!black}{($\uparrow 9.9$)} \\

\midrule

\rowcolor{gray!7}
\multicolumn{4}{l}{\textit{Mapped-Host Cosine Score}} \\[1pt]

\rowcolor{white}
Low Matched
& 0.904
& 0.920
& 0.967 \\

Medium Matched
& \underline{0.921}
& \underline{0.930}
& \underline{0.970} \\

\rowcolor{blue!6}
\textbf{High Matched}
& \textbf{0.947}
& \textbf{0.939}
& \textbf{0.972} \\

\bottomrule
\end{tabular}%
}

\vspace{-1em}
\end{table}

\noindent\textbf{Alignment and Transfer.} To investigate the relationship between representation alignment quality and cross-model capability transfer, we rank candidate source and host layer pairs according to the Geometry Match score defined in Eq.~\ref{eq:superai_geometry_match}. We select layer pairs from the top, middle, and bottom of the ranking as high, medium, and low matching configurations, respectively. As shown in Table~\ref{tab:layer_pair_quality}, downstream performance consistently improves with higher matching levels across all three host models. Specifically, low matching layer pairs improve Llama, Qwen, and Pythia by 4.4, 2.8, and 5.5 points, respectively. The gains increase to 6.8, 4.2, and 8.7 points for medium matching pairs, and further to 7.5, 5.1, and 9.9 points for high matching pairs. Consistent with this trend, the mapped-host cosine similarity also increases with matching quality, indicating that higher matching layer pairs generally produce more similar source and host representations after mapping. Overall, these results show a consistent positive relationship between geometric correspondence and downstream transfer performance among the evaluated layer pairs. Using geometric alignment information to guide the selection of injection locations can therefore improve the effectiveness of cross-architecture capability transfer.

\begin{figure}[t]
    \centering
    \includegraphics[width=0.98\linewidth]{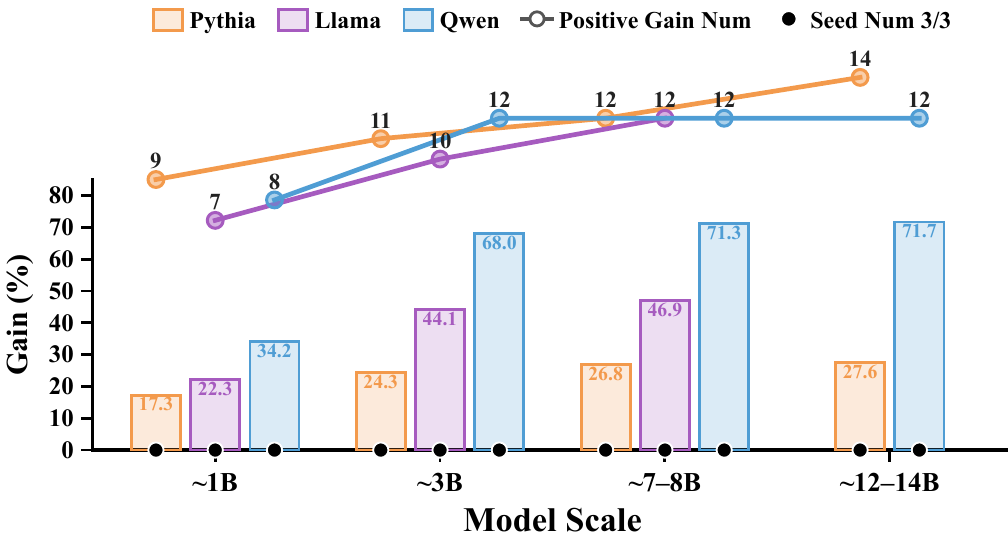}
    \vspace{-0.3cm}
\caption{\textbf{Effect of Host Model Scale.}
Bars show the Gain Rate computed from 15-domain macro-average scores.
Lines show the positive-transfer coverage $k_{+}$, defined as the number of domains with a positive performance change.
Labels of 3/3 indicate that the 15-domain average gain is positive in all three random seeds. Missing entries indicate unavailable model scales.}
\vspace{-0.3cm}
    \label{fig:numbergain}
\end{figure}

\noindent\textbf{Host Model Scale.} 
To investigate the effect of host model scale on multi-source capability fusion, we integrate the same 15 source models into Qwen, Llama, and Pythia models of different scales. Gain Rate measures the fraction of the positive source--host gap closed by fusion using 15-domain average scores. The detailed definition is provided in the appendix. As shown in Figure~\ref{fig:numbergain}, positive-transfer coverage generally increases with host model scale. At approximately 1B, Qwen, Llama, and Pythia achieve positive gains in 8, 7, and 9 domains, respectively. At the 7–8B scale, all three model families reach 12 domains, while Pythia further reaches 14 domains at the 12–14B scale. The Gain Rate also generally increases with model scale across all three families, with faster growth at smaller scales followed by slower growth at larger scales. Specifically, the Gain Rates of Qwen, Llama, and Pythia increase from 34.2\%, 22.3\%, and 17.3\% to 71.7\%, 46.9\%, and 27.6\% at their largest evaluated scales, respectively. As host models become larger, their original capabilities improve, reducing the performance gap with source models and limiting further improvements. Positive average gains across all three random seeds indicate stable capability transfer. Table~\ref{tab:qwen_domain_scaling} further examines domain-wise capability absorption across Qwen2.5 scales. Math gains peak at 19.4 points for 3B, then decline to 14.4 and 2.3 points for 7B and 14B, respectively. Physics gains decrease consistently with model scale, while Law exhibits negative transfer at 14B, with a 5.6-point performance drop. The four-domain average gain peaks at 14.7 points for 3B, then falls to 7.6 and 1.7 points for 7B and 14B, respectively. These results show that larger hosts do not necessarily achieve greater gains, suggesting that capability absorption also depends on the host's existing capabilities and remaining room for improvement.

\begin{table}
\centering
\caption{\textbf{Domain-Wise Capability Absorption across Qwen2.5 Scales.}
Each cell reports the post-fusion score and its change from the base model. Avg. denotes the four-domain average.}
\vspace{-0.3cm}
\label{tab:qwen_domain_scaling}
\scriptsize
\setlength{\tabcolsep}{2.4pt}
\renewcommand{\arraystretch}{1.10}

\resizebox{\linewidth}{!}{%
\begin{tabular}{@{}lccccc@{}}
\toprule
\textbf{Host Model}
& \textbf{Math}
& \textbf{Safety}
& \textbf{Law}
& \textbf{Physics}
& \textbf{Avg.} \\
\midrule

Qwen2.5-1.5B~\cite{qwen2.5}
& 14.3 \textcolor{red!65!black}{($\downarrow 2.1$)}
& 72.5 \textcolor{green!45!black}{($\uparrow 8.0$)}
& 49.3 \textcolor{green!45!black}{($\uparrow 8.5$)}
& 94.0 \textcolor{green!45!black}{($\uparrow 14.1$)}
& 57.5 \textcolor{green!45!black}{($\uparrow 7.1$)} \\

\addlinespace[2pt]

Qwen2.5-3B~\cite{qwen2.5}
& 45.7 \textcolor{green!45!black}{($\uparrow 19.4$)}
& 76.6 \textcolor{green!45!black}{($\uparrow 15.3$)}
& 53.5 \textcolor{green!45!black}{($\uparrow 14.1$)}
& 94.9 \textcolor{green!45!black}{($\uparrow 9.8$)}
& 67.7 \textcolor{green!45!black}{($\uparrow 14.7$)} \\

\addlinespace[2pt]

Qwen2.5-7B~\cite{qwen2.5}
& \underline{59.3} \textcolor{green!45!black}{($\uparrow 14.4$)}
& \underline{79.3} \textcolor{green!45!black}{($\uparrow 3.3$)}
& \underline{56.3} \textcolor{green!45!black}{($\uparrow 8.5$)}
& \underline{95.7} \textcolor{green!45!black}{($\uparrow 4.3$)}
& \underline{72.7} \textcolor{green!45!black}{($\uparrow 7.6$)} \\

\addlinespace[2pt]

Qwen2.5-14B~\cite{qwen2.5}
& \textbf{76.7} \textcolor{green!45!black}{($\uparrow 2.3$)}
& \textbf{80.2} \textcolor{green!45!black}{($\uparrow 8.9$)}
& \textbf{59.2} \textcolor{red!65!black}{($\downarrow 5.6$)}
& \textbf{97.9} \textcolor{green!45!black}{($\uparrow 1.1$)}
& \textbf{78.5} \textcolor{green!45!black}{($\uparrow 1.7$)} \\

\bottomrule
\end{tabular}
}
\vspace{-0.3cm}
\end{table}

\section{Conclusion}
In this work, we introduced SuperAI, a framework for integrating multiple heterogeneous source models into a single host model through orthogonal subspace alignment and lightweight residual fusion. Our analysis reveals that task-relevant representations across different model architectures exhibit approximately orthogonal correspondences in low-dimensional functional subspaces, providing a shared geometric interface for cross-model capability transfer. Experiments across 15 capability domains and three host model families demonstrate that SuperAI consistently outperforms existing cross-architecture merging methods while better preserving the host model's general capabilities. Further results across unseen datasets, model architectures, and parameter scales confirm the generalizability of this approach, while the scaling analysis shows that capability absorption is jointly constrained by model capacity and representational characteristics. Overall, our findings suggest that aligning and composing functional representations offers a promising alternative to parameter-level merging for modular and efficient multi-source integration.

\bibliographystyle{plainnat}
\bibliography{main}

\clearpage

\beginappendix
This appendix provides additional details, analyses, and discussions that complement the main paper.
Appendix~\ref{app:source_models_datasets_metrics} describes the domain-specific source models, training datasets, evaluation benchmarks, and metrics used across the 15 capability domains.
Appendix~\ref{app:notation} summarizes the notation used throughout SuperAI.
Appendix~\ref{more_experiment} presents qualitative examples and additional experiments on PCA subspace dimensionality, training-data scale, multi-source scalability, residual scaling, and source-model contributions. 
Appendix~\ref{app:shared_layer_selection} provides further details on cross-layer matching and the selection of the shared host layer.
Appendix~\ref{lim} discusses the current scope and limitations of our evaluation, together with directions for future work.
Finally, Appendix~\ref{dis} provides additional discussion on several methodological and experimental questions concerning SuperAI.

\section{Domain-Specific Source Models, Datasets, and Metrics}
\label{app:source_models_datasets_metrics}
\subsection{Source Model and Dataset}
We describe the source model, merging-training data, and
evaluation metric for each of the 15 capability domains.
Training datasets refer to the data used in our merging
procedure, rather than necessarily the original fine-tuning
corpora of the source models. All reported metrics are
higher-is-better.

\medskip
\noindent\textbf{Science.}
We use ScienceQA-OFT-8B~\cite{yang2026orthogonal} as the source model and
ARC-Challenge~\cite{clark2018think} for merging training and evaluation.
The task consists of multiple-choice science questions.
We report accuracy, measured by whether the predicted
option matches the reference answer.

\medskip
\noindent\textbf{Commonsense.}
We use CommonsenseQA-OFT-8B~\cite{yang2026orthogonal} as the source model
and CommonsenseQA~\cite{talmor2019commonsenseqa} as the corresponding dataset.
The task requires selecting the correct answer to questions
involving everyday commonsense knowledge.
The primary metric is multiple-choice accuracy.

\medskip
\noindent\textbf{Social.}
We use Social-IQA-OFT-8B~\cite{yang2026orthogonal} as the source model
and the BIG-bench Social IQA~\cite{sap2019social} task for merging training
and evaluation.
Questions concern people's intentions, reactions, and
likely consequences in social situations.
We report multiple-choice accuracy.

\medskip
\noindent\textbf{Math.}
We use NuminaMath-OFT-8B~\cite{yang2026orthogonal} as the source model
and SVAMP~\cite{patel2021nlp} for merging training and evaluation.
The task requires generating numeric answers to
arithmetic word problems.
We extract the final numeric value from the generated
response and compare it with the reference using relative
and absolute tolerances of $10^{-4}$.
The primary metric is numeric answer accuracy.

\medskip
\noindent\textbf{Code.}
We use Qwen2.5-Coder-7B-Instruct~\cite{hui2024qwen2} as the source model
and MBPP~\cite{austin2021program} for merging training.
We report Pass@1: a task is counted as solved only
when a single generated program passes both the base
and additional tests.

\medskip
\noindent\textbf{Multilingual.}
We use MergeBench Llama-3.2-3B Multilingual~\cite{he2026mergebench} as the
source model and an eight-language subset of M-ARC~\cite{clark2018think}.
The task evaluates multiple-choice reasoning across
languages.
We report sample-averaged accuracy, pooling correct
predictions over all evaluated examples rather than
taking an unweighted average of per-language accuracies.

\begin{table*}[t]
\captionsetup{font=footnotesize,skip=3pt}
\caption{\textbf{Training datasets and evaluation metrics for the 15 capability domains.}
We list the source models, training datasets, and corresponding evaluation metrics. All metrics are higher-is-better.}
\label{tab:expert_model_training}
\vspace{-0.15cm}
\centering

\footnotesize
\setlength{\tabcolsep}{4pt}
\renewcommand{\arraystretch}{1.10}

\resizebox{\textwidth}{!}{%
\begin{tabular}{llll}
\toprule
\textbf{Domain}
& \textbf{Source Model}
& \textbf{Training Dataset}
& \textbf{Task Metric} $\uparrow$ \\
\midrule

\rowcolor{gray!15}
Science
& ScienceQA-OFT-8B~\cite{yang2026orthogonal}
& ARC-Challenge~\cite{clark2018think}
& Accuracy \\

Commonsense
& CommonsenseQA-OFT-8B~\cite{yang2026orthogonal}
& CommonsenseQA~\cite{talmor2019commonsenseqa}
& Accuracy \\

\rowcolor{gray!15}
Social
& Social-IQA-OFT-8B~\cite{yang2026orthogonal}
& BIG-bench Social IQA ~\cite{sap2019social}
& Accuracy \\

Math
& NuminaMath-OFT-8B~\cite{yang2026orthogonal}
& SVAMP~\cite{patel2021nlp}
& Accuracy \\

\rowcolor{gray!15}
Code
& Qwen2.5-Coder-7B-Instruct~\cite{hui2024qwen2}
& MBPP~\cite{austin2021program}
& Pass@1 \\

Multilingual
& Llama-3.2-3B Multilingual~\cite{he2026mergebench}
& M-ARC~\cite{clark2018think}
& Averaged Accuracy \\

\rowcolor{gray!15}
Safety
& Llama-3.2-3B Safety~\cite{he2026mergebench}
& PKU-SafeRLHF~\cite{ji2025pku}
& Accuracy \\

Law
& Saul-Instruct-v1~\cite{colombo2024saullm}
& MMLU Law~\cite{hendrycks2020measuring}
& Accuracy \\

\rowcolor{gray!15}
Finance/Business
& Finance Llama-3.1-8B~\cite{finance_llama_3.1_8b}
& FinQA~\cite{chen2022finqadatasetnumericalreasoning}
& Accuracy \\

Astronomy
& AstroSage-8B~\cite{de_Haan_2025}
& Astro-ph~\cite{lambrides2024uncovering}
& Accuracy \\

\rowcolor{gray!15}
Medical
& HuatuoGPT-o1-7B~\cite{chen2024huatuogpto1medicalcomplexreasoning}
& MedMCQA~\cite{pal2022medmcqalargescalemultisubject}
& Accuracy \\

Climate
& ClimateGPT-7B~\cite{thulke2024climategptaisynthesizinginterdisciplinary}
& Climate-FEVER Binary~\cite{diggelmann2021climatefeverdatasetverificationrealworld}
& Accuracy \\

\rowcolor{gray!15}
Cybersecurity
& Foundation-Sec-8B~\cite{kassianik2025llama31foundationaisecurityllmbase8btechnicalreport}
& SEVENLLM~\cite{ji2024sevenllmbenchmarkingelicitingenhancing}
& F1 \\

Chemistry
& ChemDFM-v1.5-8B~\cite{zhao2024chemdfm}
& MMLU Chemistry~\cite{hendrycks2020measuring}
& Accuracy \\

\rowcolor{gray!15}
Physics
& FeynTune-S1~\cite{richmond2026feyntune}
& Hep-th Continuation~\cite{richmond2026feyntune}
& Accuracy \\

\bottomrule
\end{tabular}%
}

\vspace{-0.3cm}
\end{table*}

\medskip
\noindent\textbf{Safety.}
We use MergeBench Llama-3.2-3B Safety~\cite{he2026mergebench} as the source model
and PKU-SafeRLHF~\cite{ji2025pku} as the corresponding dataset.
The task presents a user request and candidate responses,
and asks the model to select the safer response.
We report safer-response selection accuracy.
This metric evaluates safety-related response selection
rather than the safety of freely generated responses.

\medskip
\noindent\textbf{Law.}
We use Saul-Instruct-v~\cite{colombo2024saullm} as the source model.
Merging training uses the local MMLU Law
subset~\cite{hendrycks2020measuring}.
Both are formulated as multiple-choice legal questions.
The primary metric is answer-selection accuracy.

\medskip
\noindent\textbf{Finance/Business.}
We use Finance Llama-3.1-8B~\cite{finance_llama_3.1_8b} as the source model
and FinQA~\cite{chen2022finqadatasetnumericalreasoning} for merging training and evaluation.
The task requires answering questions based on
financial information.
Numeric answers are extracted and compared using
relative and absolute tolerances of $10^{-4}$.
For nonnumeric references, we use case-insensitive
reference-answer matching in the generated response.
We report the fraction of answers judged correct
under these rules.

\medskip
\noindent\textbf{Astronomy.}
We use AstroSage-8B~\cite{de_Haan_2025} as the source model
and arXiv astro-ph~\cite{richmond2026feyntune} text for merging training.
Evaluation uses MMLU Astronomy~\cite{hendrycks2020measuring} multiple-choice questions.
The primary metric is accuracy on MMLU Astronomy;
this score is distinct from language-modeling evaluation
on the arXiv training corpus.

\medskip
\noindent\textbf{Medical.}
We use HuatuoGPT-o1-7B~\cite{chen2024huatuogpto1medicalcomplexreasoning} as the source model
and MedMCQA~\cite{pal2022medmcqalargescalemultisubject} for merging training and evaluation.
The task consists of multiple-choice questions
covering medical knowledge.
We report accuracy based on agreement with the
reference answer option.

\medskip
\noindent\textbf{Climate.}
We use ClimateGPT-7B~\cite{thulke2024climategptaisynthesizinginterdisciplinary} as the source model
and the local binary formulation of Climate-FEVER~\cite{diggelmann2021climatefeverdatasetverificationrealworld}.
The task requires classifying climate-related statements
according to the provided binary labels.
The primary metric is classification accuracy.

\medskip
\noindent\textbf{Cybersecurity.}
We use Foundation-Sec-8B~\cite{kassianik2025llama31foundationaisecurityllmbase8btechnicalreport} as the source model
and SEVENLLM~\cite{ji2024sevenllmbenchmarkingelicitingenhancing} for merging training and evaluation.
The task requires generating responses to
cybersecurity-related instructions.
We report token-level F1 between generated and reference
responses, combining token precision and recall.
Scores are computed for individual examples and
then averaged over the evaluation set.

\medskip
\noindent\textbf{Chemistry.}
We use ChemDFM-v1.5-8B~\cite{zhao2024chemdfm} as the source model.
Merging training uses a local mixture of the MMLU~\cite{hendrycks2020measuring}
High School Chemistry and College Chemistry subsets,
containing 165 and 6 training examples, respectively.
Evaluation uses 100 questions from MMLU~\cite{hendrycks2020measuring} College Chemistry.
The primary metric is multiple-choice accuracy.

\medskip
\noindent\textbf{Physics.}
We use FeynTune-S1 Llama-3.1-8B~\cite{richmond2026feyntune} as the source model
and continuation data constructed from arXiv hep-th~\cite{richmond2026feyntune} text.
The task requires selecting the true next sentence
from a set of candidates.
We report next-sentence selection accuracy, measuring
whether the selected continuation matches the reference.

\medskip
\noindent\textbf{Aggregate Evaluation.}
For multi-source fusion, we report the unweighted mean
of the 15 domain scores after expressing them on a
common scale. Each domain receives equal weight,
regardless of its evaluation-set size.
This aggregate is a mean of task-specific metrics,
rather than a pooled accuracy over all questions.
We additionally report performance on a separate
MMLU evaluation set to assess general-capability
retention; this score is excluded from the
15-domain average.

\medskip
\noindent\textbf{Generalization Evaluation.}
For cross-dataset generalization, we evaluate Science, Commonsense, Social, and Climate capabilities on OpenBookQA~\cite{mihaylov2018suitarmorconductelectricity}, PIQA~\cite{bisk2019piqareasoningphysicalcommonsense}, CosmosQA~\cite{huang2019cosmosqamachinereading}, and ClimateCheck~\cite{abu-ahmad-etal-2025-climatecheck}, respectively, none of which is used for capability merging or residual-MLP training.

\subsection{Metrics}
\noindent\textbf{Gain Rate.} Let $D=15$ denote the number of capability domains. For host model $m$ and domain $d$, let $B_{m,d}$ denote the score of the original host model, $S_d$ the score of the corresponding
domain-specific source model, and $F_{m,d}$ the post-fusion score of SuperAI. We first define the average positive source--host gap as
\begin{equation}
\Delta^{+}_{\mathrm{src},m}
=
\frac{1}{D}
\sum_{d=1}^{D}
\max\!\left(S_d-B_{m,d},0\right).
\end{equation}
This quantity measures the average amount of domain capability
available for transfer from the source models to host model $m$. A domain contributes zero to this quantity when its source model does not outperform the original host. The average gain achieved by fusion is
\begin{equation}
\Delta_{\mathrm{fuse},m}
=
\frac{1}{D}
\sum_{d=1}^{D}
\left(F_{m,d}-B_{m,d}\right).
\end{equation}
We define the Gain Rate as the fraction of the positive
source--host gap closed by fusion:
\begin{equation}
\operatorname{GainRate}_{m}
=
100
\times
\frac{
\Delta_{\mathrm{fuse},m}
}{
\Delta^{+}_{\mathrm{src},m}
}.
\end{equation}
The Gain Rate is reported as a percentage. We do not clip the
numerator or the final ratio; therefore, negative transfer can
produce a negative Gain Rate, while a value above $100\%$ indicates that the average fusion gain exceeds the measured positive source--host gap. If
$\Delta^{+}_{\mathrm{src},m}=0$, the Gain Rate is undefined and the corresponding entry is reported as unavailable rather than as zero. We additionally define positive-transfer coverage as
\begin{equation}
k_{+,m}
=
\sum_{d=1}^{D}
\mathbb{I}\!\left[F_{m,d}>B_{m,d}\right],
\end{equation}
where $\mathbb{I}[\cdot]$ is the indicator function. Thus,
$k_{+,m}$ is the number of domains in which the fused model
outperforms its original host, and its value ranges from $0$ to $15$. For experiments with three random seeds, a label of $3/3$ indicates that the 15-domain macro-average gain,
\begin{equation}
\frac{1}{D}
\sum_{d=1}^{D}
\left(F^{(p)}_{m,d}-B_{m,d}\right),
\end{equation}
is positive for each of the three seeds $p$. Missing entries in the scale analysis indicate unavailable host-model scales and are not treated as zero-valued results. All averages and Gain Rates are computed from unrounded domain-level scores and rounded only for presentation.

\noindent\textbf{Mapped--Host Cosine Score.} For an evaluation prompt $a_i$ associated with source $s_i$, let
$\widetilde{\mathbf{h}}_{s_i\rightarrow h}(a_i)$ denote the mapped source representation in the host hidden space, and let
$\mathbf{h}'_{i,p_i}$ denote the host representation at the last non-padding prompt position after injecting the MLP-predicted residual. We define the mapped--host cosine score as
\begin{equation}
C_{\mathrm{map}}
=
\frac{1}{N_{\mathrm{eval}}}
\sum_{i=1}^{N_{\mathrm{eval}}}
\frac{
\left\langle
\widetilde{\mathbf{h}}_{s_i\rightarrow h}(a_i),
\mathbf{h}'_{i,p_i}
\right\rangle
}{
\left\|
\widetilde{\mathbf{h}}_{s_i\rightarrow h}(a_i)
\right\|_2
\left\|
\mathbf{h}'_{i,p_i}
\right\|_2
}.
\end{equation}
A higher value indicates that the post-injection host
representation is more closely aligned with the mapped source representation.

\subsection{Implementation Details}
\noindent\textbf{Data Splits. }For each capability domain, we use a fixed data-splitting protocol
shared across all three host models. The independent test set is reserved exclusively for final evaluation. The remaining non-test data are divided into approximately 80\% for training and alignment fitting and 20\% for validation. The training and alignment-fitting split is used to estimate the normalization statistics, fit the PCA projections and orthogonal Procrustes mappings, and train the residual adapter. The validation split is used for source--host layer selection, hyperparameter selection, and checkpoint selection. The test split is not used for PCA fitting, Procrustes alignment, residual-adapter training, layer selection, or any other form of hyperparameter selection. The training, validation, and test splits contain no overlapping examples. The same fixed splits are used for all host models, comparison methods, and random seeds.

\noindent\textbf{Training and Evaluation Settings. }For all three host models, the backbone parameters are frozen and
only the lightweight residual adapter is trained. The joint
training objective is
\begin{equation}
\mathcal{L}
=
\lambda_{\mathrm{task}}\mathcal{L}_{\mathrm{task}}
+
\lambda_{\mathrm{inc}}\mathcal{L}_{\mathrm{inc}},
\end{equation}
where $\mathcal{L}_{\mathrm{task}}$ is the answer-token
cross-entropy loss and $\mathcal{L}_{\mathrm{inc}}$ is the
source-derived increment-supervision loss. We set
$\lambda_{\mathrm{task}}=1$ and $\lambda_{\mathrm{inc}}=0.5$. The residual scaling factor is set to $\alpha=1.0$. We use a
128-dimensional PCA subspace and an adapter bottleneck dimension of 128. The residual adapter is optimized using AdamW with a learning rate of $2\times10^{-4}$, a weight decay of $0.01$, and a batch size of 4 for 2 epochs. Each experiment is independently run three times using random seeds 42, 44, and 46, and we report the average results across the three runs.

\section{Notation Summary}
\label{app:notation}

\begin{table}[!t]
\centering
\caption{Summary of Notation Used in SuperAI (Part I).}
\label{tab:notation}
\footnotesize
\renewcommand{\arraystretch}{1.08}
\begin{tabular}{@{}p{0.34\linewidth}p{0.62\linewidth}@{}}
\toprule
Symbol & Definition \\
\midrule
\multicolumn{2}{@{}l}{\textit{Models and Representations}}\\

$f_h,\{f_s\}_{s=1}^{K}$
& Host model and $K$ source models. \\

$d_h,d_s$
& Hidden dimensions of the host and source model $s$. \\

$\mathcal{A}_s=\{a_i\}_{i=1}^{n_s}$
& Shared anchor prompts for source model $s$ and the host. \\

$\mathcal{A}_s^{\mathrm{fit}}$
& Alignment-fitting prompt set. \\

$V_s$
& Validation set used for layer selection. \\

$\ell_m,h_m^{(\ell_m)}(a_i)$
& Layer index and last non-padding-token hidden state of model $m\in\{s,h\}$. \\

$H_m^{(\ell_m)}$
& Matrix of hidden states for prompts in $\mathcal{A}_s$. \\

$\mathcal{L}_h,\mathcal{L}_s$
& Candidate layer sets of the host and source models. \\

\midrule
\multicolumn{2}{@{}l}{\textit{Subspace Alignment}}\\

$\mu_m,u_m(a_i),U_m$
& Hidden-state mean, centered and normalized state, and matrix of normalized states. \\

$c_m,P_m,r$
& PCA mean, principal-component matrix, and shared subspace dimension. \\

$I_r$
& Identity matrix of dimension $r$. \\

$Z_m,z_m(a_i)$
& PCA coordinate matrix and coordinate vector of model $m$. \\

$Z_{m,\mathrm{cal}},
Z_{m,\mathrm{val}}$
& PCA coordinates of calibration and validation samples. \\

$L,\Sigma,V$
& Matrices in the SVD used to compute the orthogonal mapping. \\

$R_s^*$
& Optimal orthogonal source-to-host mapping. \\

$\widehat Z_{s\to h},
\widehat z_{s\to h}(a_i)$
& Aligned source-coordinate matrix and vector. \\

\midrule
\multicolumn{2}{@{}l}{\textit{Layer Matching and Selection}}\\

$C_{\mathrm{pair},s}$
& Mean paired cosine similarity between aligned source and host coordinates. \\

$C_{\mathrm{CKA},s}$
& Linear CKA between source and host PCA coordinate matrices. \\

$E_{\mathrm{nrmse},s}$
& Normalized reconstruction error after alignment. \\

$S_s(\ell_s,\ell_h)$
& Geometry Match score for a source--host layer pair. \\

$\ell_s^*(\ell_h)$
& Best source layer for a candidate host layer. \\

$\ell_h^*,\ell_s^*$
& Selected shared host layer and final source-specific layer. \\

\bottomrule
\end{tabular}
\end{table}

\begin{table}[!t]
\centering
\caption{Summary of notation used in SuperAI (Part II).}
\label{tab:notation2}
\footnotesize
\renewcommand{\arraystretch}{1.08}
\begin{tabular}{@{}p{0.34\linewidth}p{0.62\linewidth}@{}}
\toprule
Symbol & Definition \\
\midrule
\multicolumn{2}{@{}l}{\textit{Feature Mapping}}\\

$z_s(a_i),
\widehat z_{s\to h}(a_i)$
& Source PCA coordinates and their rotated host-space coordinates. \\

$b_{s\to h}(a_i)$
& Rotated coordinates reconstructed in the normalized host hidden space. \\

$\widetilde h_{s\to h}(a_i)$
& Mapped host-space representation after magnitude restoration. \\

$h_h(a_i)$
& Original host representation at the selected host layer. \\

$\Delta_s(a_i)$
& Target feature increment
$\widetilde h_{s\to h}(a_i)-h_h(a_i)$. \\

\midrule
\multicolumn{2}{@{}l}{\textit{Residual Fusion and Optimization}}\\

$h_{i,t},h'_{i,t}$
& Host hidden state before and after residual injection at token position $t$. \\

$b,W_{\mathrm{down}},
W_{\mathrm{up}}$
& MLP bottleneck width, down-projection, and up-projection. \\

$\mathrm{LN},\mathrm{SiLU}$
& Layer normalization and SiLU activation function. \\

$q_{i,t},
\widehat\Delta_{i,t}$
& Bottleneck activation and predicted residual increment. \\

$\alpha$
& Residual scaling factor controlling the injected update strength. \\

$\mathcal{A}^{\mathrm{tr}}_s,N$
& Source-$s$ training prompts and total training-prompt count. \\

$s_i,p_i$
& Source identity of prompt $a_i$ and its last non-padding position. \\

$\rho_s$
& Source-specific increment normalization factor. \\

$T_i,y_i,y_{i,t}$
& Answer-token positions, ground-truth answer, and answer token. \\

$p_\theta,\theta$
& Model output distribution and trainable MLP parameters. \\

$\mathcal{L}_{\mathrm{task}}$
& Answer-token cross-entropy loss. \\

$\mathcal{L}_{\mathrm{inc}}$
& Source-derived increment supervision loss. \\

$\lambda_{\mathrm{task}},
\lambda_{\mathrm{inc}}$
& Weights of task and increment supervision losses. \\

$\mathcal{L}$
& Weighted joint training objective. \\

$\epsilon$
& Positive constant used for numerical stability. \\

\bottomrule
\end{tabular}
\end{table}

\begin{table}[!t]
\centering
\caption{Summary of notation used in SuperAI (Part III).}
\label{tab:notation3}
\footnotesize
\renewcommand{\arraystretch}{1.08}
\begin{tabular}{@{}p{0.34\linewidth}p{0.62\linewidth}@{}}
\toprule
Symbol & Definition \\
\midrule
\multicolumn{2}{@{}l}{\textit{Evaluation Metrics}}\\

$D,d$
& Number of capability domains and domain index. \\

$B_{m,d}$
& Original host model score on domain $d$. \\

$S_d$
& Domain-specific source model score on domain $d$. \\

$F_{m,d}$
& Post-merging score of host model $m$ on domain $d$. \\

$\Delta_{\mathrm{src},m}^{+}$
& Average positive source--host performance gap. \\

$\Delta_{\mathrm{fuse},m}$
& Average performance gain after merging. \\

$\mathrm{GainRate}_m$
& Percentage of the positive source--host gap closed by merging. \\

$k_{+,m}$
& Number of domains with positive performance gains. \\

$\mathbb{I}[\cdot]$
& Indicator function for positive-transfer coverage. \\

$F_{m,d}^{(p)}$
& Post-merging domain score under random seed $p$. \\

$C_{\mathrm{map}}$
& Mean cosine similarity between mapped source and post-injection host representations. \\

$N_{\mathrm{eval}}$
& Number of evaluation prompts used for $C_{\mathrm{map}}$. \\

\midrule
\multicolumn{2}{@{}l}{\textit{Experimental Configurations}}\\

$\mathcal{O}_0$
& Fixed source ordering for overall capability and MMLU evaluation. \\

$\mathcal{O}_d$
& Target-first source ordering for domain-specific scalability evaluation. \\

$d$
& Target domain used to construct the source ordering. \\

$\parallel$
& Ordered concatenation operator. \\

\bottomrule
\end{tabular}
\end{table}

Tables~\ref{tab:notation}, \ref{tab:notation2}, and \ref{tab:notation3}
summarize the notation used throughout SuperAI. Table~\ref{tab:notation} covers model representations, orthogonal subspace alignment, and layer selection. Table~\ref{tab:notation2} presents the symbols for feature mapping, residual injection, and joint optimization. Table~\ref{tab:notation3} defines the evaluation metrics and experimental configurations. We use $h$ and $s$ to denote the host model and a source model, respectively, while $\ell_h$ and $\ell_s$ index their corresponding layers. Shared anchor prompts are used to extract paired hidden representations from the source and host models. These representations are normalized and projected into low dimensional PCA subspaces. The resulting coordinates are aligned through an orthogonal mapping, and the Geometry Match scores are used for source--host layer matching and shared host-layer selection.

After alignment, the source representations are mapped into the host hidden space to construct source-specific target feature increments, denoted by $\Delta_s(a_i)$. These increments provide representation-level supervision for the MLP. During joint training, the MLP predicts input-dependent residual updates from the host hidden states. The predicted updates are injected into the selected host layer. The training objective combines task-output supervision and source-increment supervision using the corresponding loss weights.

For each source model, the host-side normalization statistics, PCA mean, and PCA basis are fitted using the anchor prompts paired with that source model. Therefore, these quantities are source-dependent even when they belong to the host representation space. For simplicity, we omit the source index when its meaning is clear from the context. The same convention applies to other source-dependent intermediate quantities.

\section{More Experiments}
\label{more_experiment}

\begin{figure*}
    \centering
    \includegraphics[width=0.98\linewidth]{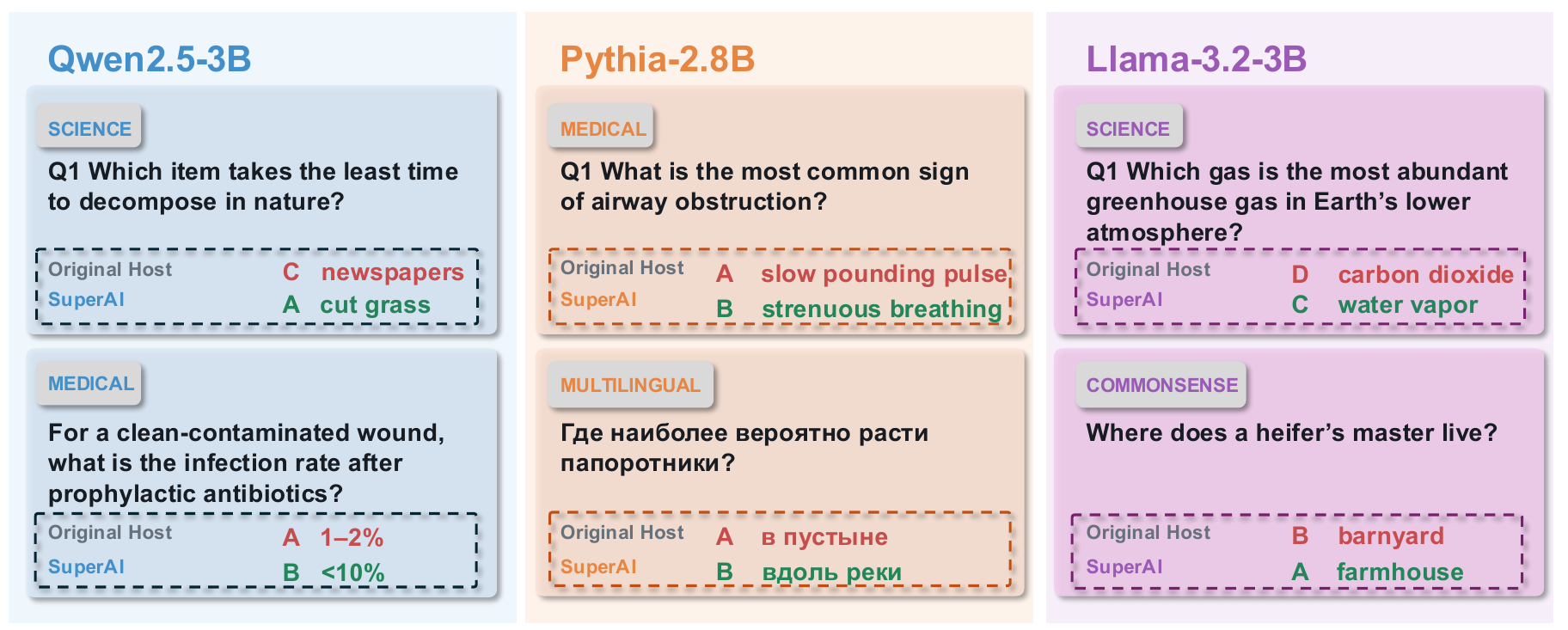}
    \vspace{-0.3cm}
\caption{\textbf{Qualitative Examples of Capability Improvements Achieved by SuperAI.} Six representative cases across three host models illustrate how SuperAI corrects original host predictions in different capability domains.}
    \label{fig:case}
\end{figure*}

\subsection{Qualitative Analysis}
To provide a more intuitive assessment of how SuperAI improves the host models, we present six representative cases in which the original host produces an incorrect answer while the fused model answers correctly. As shown in the figure, SuperAI corrects Qwen2.5-3B from “newspapers” to “cut grass” on a science question and from “1–2\%” to “<10\%” on a medical question. For Pythia-2.8B, SuperAI correctly identifies “strenuous breathing” as the most common sign of airway obstruction and changes the answer to the Russian question about fern habitats from “desert” to “along a river.” For Llama-3.2-3B, SuperAI corrects “carbon dioxide” to “water vapor” as the most abundant greenhouse gas in the lower atmosphere and changes “barnyard” to “farmhouse” in a commonsense question. Covering science, medicine, multilingual understanding, and commonsense reasoning, these examples illustrate how SuperAI corrects different types of errors made by heterogeneous host models by incorporating complementary knowledge from the source models. This qualitative analysis complements the aggregate quantitative results rather than serving as standalone evidence of general performance improvements.

\subsection{Effect of PCA Subspace Dimension}

\begin{table}
\vspace{-1.2em}
\centering
\caption{
\textbf{Effect of PCA Subspace Dimensionality.}
Each cell reports the 15-domain average score, with the absolute
change from the corresponding base model shown in parentheses.
PCA rank denotes the requested upper bound.
}
\label{tab:pca_ablation_shifted}
\scriptsize
\setlength{\tabcolsep}{2.8pt}
\renewcommand{\arraystretch}{1}
\resizebox{\linewidth}{!}{
\begin{tabular}{lccc}
\toprule
\textbf{PCA Rank} & \textbf{Llama-3.2-3B} & \textbf{Qwen2.5-3B} & \textbf{Pythia-2.8B} \\
\midrule
Base & 46.7 & 56.2 & 22.9 \\
\midrule
16 & 56.3 \textcolor{green!45!black}{($\uparrow 9.6$)} & 60.2 \textcolor{green!45!black}{($\uparrow 4.0$)} & 28.7 \textcolor{green!45!black}{($\uparrow 5.8$)} \\
32 & 55.3 \textcolor{green!45!black}{($\uparrow 8.6$)} & \textbf{61.3} \textcolor{green!45!black}{\textbf{($\uparrow 5.1$)}} & 28.3 \textcolor{green!45!black}{($\uparrow 5.4$)} \\
64 & 54.5 \textcolor{green!45!black}{($\uparrow 7.8$)} & 61.2 \textcolor{green!45!black}{($\uparrow 5.0$)} & 29.0 \textcolor{green!45!black}{($\uparrow 6.1$)} \\
128 & 54.2 \textcolor{green!45!black}{($\uparrow 7.5$)} & \textbf{61.3} \textcolor{green!45!black}{\textbf{($\uparrow 5.1$)}} & \textbf{32.8} \textcolor{green!45!black}{\textbf{($\uparrow 9.9$)}} \\
256 & 55.2 \textcolor{green!45!black}{($\uparrow 8.5$)} & 60.8 \textcolor{green!45!black}{($\uparrow 4.6$)} & 28.9 \textcolor{green!45!black}{($\uparrow 6.0$)} \\
512 & \textbf{56.9} \textcolor{green!45!black}{\textbf{($\uparrow 10.2$)}} & 59.8 \textcolor{green!45!black}{($\uparrow 3.6$)} & 28.9 \textcolor{green!45!black}{($\uparrow 6.0$)} \\
\bottomrule
\end{tabular}}
\vspace{-1em}
\end{table}
To evaluate the effect of the PCA subspace dimension on multi-source capability transfer, we vary the PCA rank from 16 to 512 and compare the performance across three host models: Llama-3.2-3B, Qwen2.5-3B, and Pythia-2.8B. The results show in Table~\ref{tab:pca_ablation_shifted} that all PCA rank settings outperform their corresponding base models, demonstrating that the proposed method effectively integrates multi-domain capabilities across different subspace dimensions. Specifically, Llama-3.2-3B achieves its highest average score of 56.9 at rank 512, improving by 10.2 points over the base model. Qwen2.5-3B attains a peak score of 61.3 across ranks 32 and 128, with a 5.1-point gain. Pythia-2.8B reaches its best score of 32.8 at rank 128, yielding a 9.9-point improvement. Thus, Llama-3.2-3B, Qwen2.5-3B, and Pythia-2.8B deliver optimal performance at ranks 512, 32 and 128, respectively. Overall performance does not scale monotonically with increasing rank. A plausible reason is that higher-dimensional subspaces retain more redundant and complex signals, which complicates optimization for the mapping module and hurts capability transfer. Considering the overall performance across all three host models, rank 128 achieves a solid balance between information preservation and training difficulty, and is therefore adopted as the default configuration.

\subsection{Effect of Training Data Scale}

\begin{table}

\centering
\caption{
\textbf{Effect of Training Data Volume.}
Each cell reports the 15-domain average score, with the absolute
change from the corresponding base model shown in parentheses.
The training cap is applied per domain; domains with fewer
examples use all available training samples.
}
\label{tab:data_ablation_shifted}
\scriptsize
\setlength{\tabcolsep}{2.8pt}
\renewcommand{\arraystretch}{1.15}
\resizebox{\linewidth}{!}{
\begin{tabular}{lccc}
\toprule
\textbf{Examples/Domain} & \textbf{Llama-3.2-3B} & \textbf{Qwen2.5-3B} & \textbf{Pythia-2.8B} \\
\midrule
Base & 46.7 & 56.2 & 22.9 \\
\midrule
50 & 54.1 \textcolor{green!45!black}{($\uparrow 7.4$)} & 57.9 \textcolor{green!45!black}{($\uparrow 1.7$)} & 27.8 \textcolor{green!45!black}{($\uparrow 4.9$)} \\
100 & 56.3 \textcolor{green!45!black}{($\uparrow 9.6$)} & 59.6 \textcolor{green!45!black}{($\uparrow 3.4$)} & 27.1 \textcolor{green!45!black}{($\uparrow 4.2$)} \\
200 & \textbf{57.6} \textcolor{green!45!black}{\textbf{($\uparrow 10.9$)}} & 58.9 \textcolor{green!45!black}{($\uparrow 2.7$)} & 26.9 \textcolor{green!45!black}{($\uparrow 4.0$)} \\
500 & 55.6 \textcolor{green!45!black}{($\uparrow 8.9$)} & 60.0 \textcolor{green!45!black}{($\uparrow 3.8$)} & 27.4 \textcolor{green!45!black}{($\uparrow 4.5$)} \\
1000 & 54.2 \textcolor{green!45!black}{($\uparrow 7.5$)} & \textbf{61.3} \textcolor{green!45!black}{\textbf{($\uparrow 5.1$)}} & \textbf{32.8} \textcolor{green!45!black}{\textbf{($\uparrow 9.9$)}} \\
\bottomrule
\end{tabular}}
\end{table}

To analyze the effect of training data scale on multi-source capability transfer, we vary the maximum number of training samples per domain from 50 to 1,000 and compare performance across three host models. The results in Table~\ref{tab:data_ablation_shifted} indicate that even with only 50 samples per domain, all three models markedly outperform their respective base models, demonstrating high data efficiency. As the sample size increases, different host models exhibit distinct trajectories: Llama-3.2-3B achieves its peak average score of 57.6 with 200 samples, whereas Qwen2.5-3B and Pythia-2.8B reach their highest scores of 61.3 and 32.8 at 1,000 samples, respectively. These findings show that increasing training data generally promotes more adequate capability mappings, yet performance does not improve monotonically with data scale. An excessive number of samples may introduce redundant or complex patterns, thereby complicating optimization for the mapping module. Overall, the proposed method yields substantial improvements in data-scarce settings, while larger training sets provide additional gains for certain host models.

\subsection{Multi-Source Scalability}

\begin{figure}[t]
\vspace{-0.8em}
    \centering
    \includegraphics[width=\linewidth]{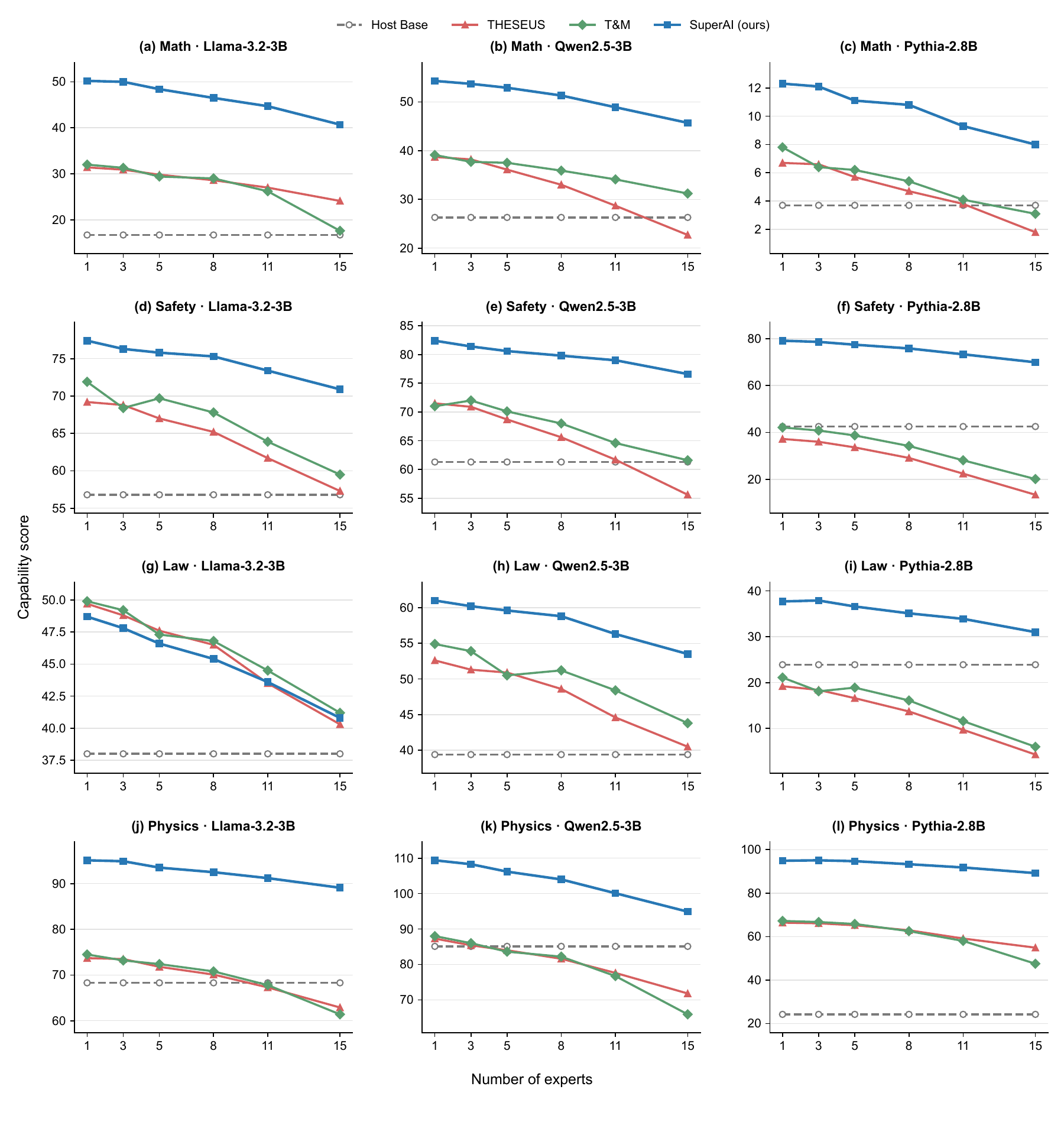}
    \vspace{-0.3cm}
    \caption{\textbf{Multi-Source Scalability across Capability Domains.} Capability scores of SuperAI, THESEUS, and T\&M as the number of merged sources increases across Math, Safety, Law, and Physics.}
    \label{fig:four_domain}
    \vspace{-0.8em}
\end{figure}

\noindent\textbf{Source Ordering.} For the overall capability and MMLU evaluation in Figure 3 of the main paper, we use the following fixed ordering of the 15 source domains:
\begin{equation}
\begin{aligned}
\mathcal{O}_0=\Bigl[&
\text{Code},\,
\text{Math},\,
\text{Science},\,
\text{Safety},
\\
&
\text{Finance/Business},\,
\text{Medical},\,
\text{Social},\,
\text{Physics},
\\
&
\text{Commonsense},\,
\text{Law},\,
\text{Chemistry},\,
\text{Astronomy},
\\
&
\text{Climate},\,
\text{Cybersecurity},\,
\text{Multilingual}
\Bigr].
\end{aligned}
\end{equation}
For each target domain $d$, we construct a target-first ordering $\mathcal{O}_d$ by moving the source model associated with $d$ to the first position while preserving the relative order of all remaining source models:
\begin{equation}
\mathcal{O}_d
=
[d]\mathbin{\Vert}
\left(\mathcal{O}_0\setminus\{d\}\right),
\end{equation}
where $\Vert$ denotes ordered concatenation. For example, the Math ordering begins with Math, Code, and
Science, whereas the Safety ordering begins with Safety, Code,
Math, and Science. The configurations with 1, 3, 5, 8, 11, and
15 sources use the corresponding prefixes of the target-specific ordering. This ensures that the source associated with the evaluated capability is included in the one-source configuration. All methods and host models use the same target-specific source subsets.

\noindent\textbf{Result. }To evaluate the scalability of cross-architecture multi-source capability fusion, we incrementally integrated 1--15 source models in a fixed order and compared SuperAI with THESEUS and Transport and Merge (T\&M) across four representative domains and three host architectures. As shown in Figure~\ref{fig:four_domain}, scores generally declined as more sources were incorporated. Averaged across the 12 domain--host combinations, SuperAI declined by 7.7 points from 1 to 15 sources, compared with 12.8 and 13.4 points for THESEUS and T\&M, respectively. SuperAI achieved the highest scores at every evaluated source count in Math, Safety, and Physics. At the 15-source endpoint, the sole domain--host combination in which it did not lead was Law on Llama-3.2-3B, where it trailed T\&M by 0.4 points. At that endpoint, SuperAI exceeded the better of the two baselines in each domain--host combination by 19.2 points on average, with particularly large margins in Safety and Physics on Pythia-2.8B and Physics on Llama-3.2-3B. These results suggest that SuperAI mitigates interference under the evaluated fixed source order, although its scores still decline as more sources are integrated.

\begin{figure}
    \centering
    \includegraphics[width=0.98\linewidth]{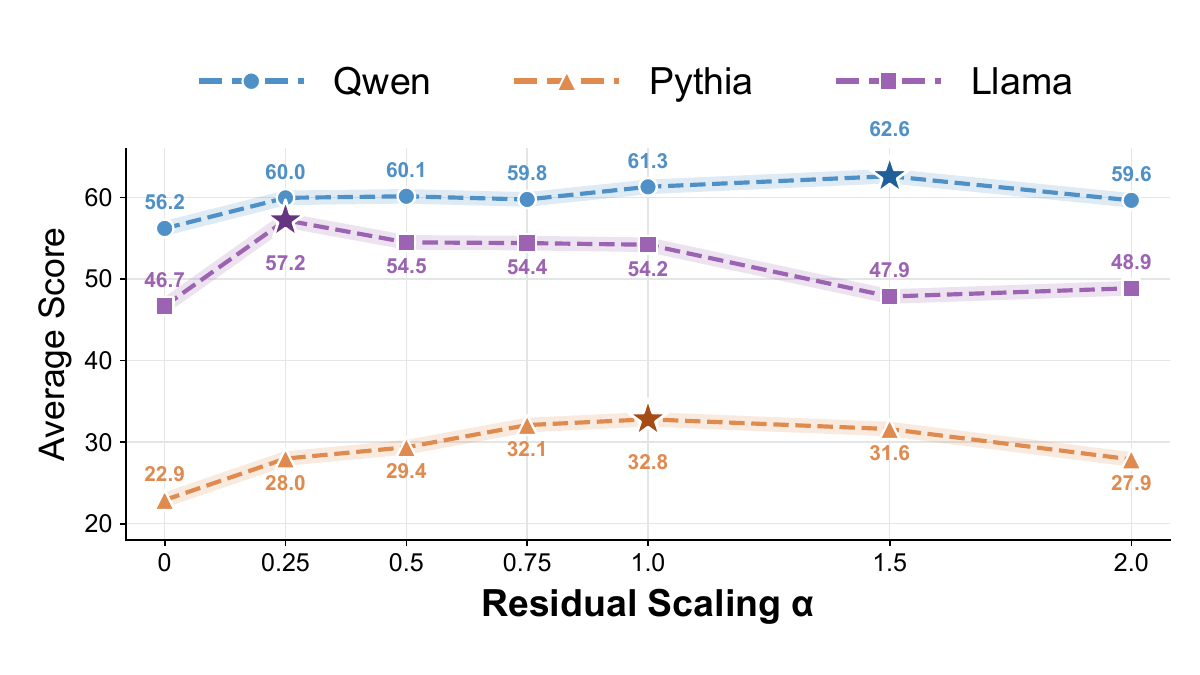}
    \vspace{-0.3cm}
\caption{\textbf{Effect of Residual Scaling Factor $\alpha$.}
Average scores across 15 capability domains for Llama-3.2-3B, Qwen2.5-3B, and Pythia-2.8B under different residual scaling factors.}
    \vspace{-0.4cm}
    \label{fig:alpha}
\end{figure}

\subsection{Effect of Residual Scaling Factor.} To investigate the effect of the residual scaling factor $\alpha$ in Eq. (10) of the main paper, we vary $\alpha$ over $\{0, 0.25, 0.5, 0.75, 1.0, 1.5, 2.0\}$ and evaluate the average performance across 15 capability domains on three host models: Llama-3.2-3B, Qwen2.5-3B, and Pythia-2.8B. As shown in Figure~\ref{fig:alpha}, the performance of all three models generally improves as $\alpha$ increases, but declines when $\alpha$ becomes too large. Specifically, Qwen2.5-3B achieves its highest score of 62.6 at $\alpha=1.5$, while Llama-3.2-3B and Pythia-2.8B reach their best scores of 57.2 and 32.8 at $\alpha=0.25$ and $\alpha=1.0$, respectively. When $\alpha=0$, all three models perform worse than their best configurations, indicating the importance of residual injection for capability transfer. However, larger values of $\alpha$ do not consistently improve performance, as overly strong residual updates may interfere with the original host representations. These results suggest that $\alpha$ controls the balance between transferring source capabilities and preserving host representations. Overall, $\alpha=1.0$ provides stable and competitive performance across all three models, and we adopt it as the default setting.

\begin{figure}
    \centering
    \includegraphics[width=0.98\linewidth]{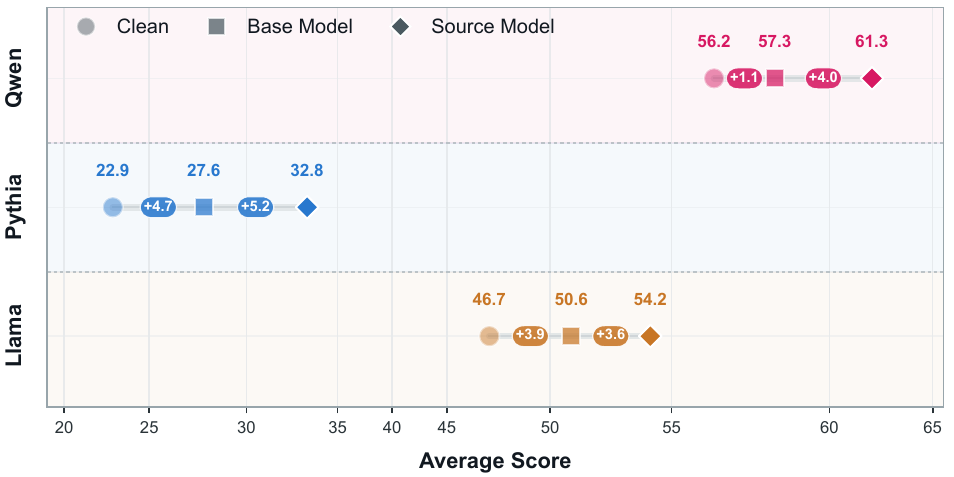}
    \vspace{-0.3cm}
\caption{\textbf{Effect of Source Model Contribution.}
Comparison of 15-domain average scores across three host models under three settings: original host (Clean), integrating base models (Base Model), and integrating domain-specific source models (Source Model).}
    \vspace{-0.4cm}
    \label{fig:hostbase}
\end{figure}

\subsection{Effect of Source Model Contribution.} To investigate whether the performance gains of SuperAI come from the specialized knowledge of expert models rather than additional training alone, we conduct an ablation study using base models. Specifically, we replace the domain-specific expert models with their corresponding base models while keeping the fusion procedure and training settings unchanged. As shown in Figure~\ref{fig:hostbase}, using base models improves the average scores of Qwen2.5-3B, Pythia-2.8B, and Llama-3.2-3B from 56.2, 22.9, and 46.7 to 57.3, 27.6, and 50.6, respectively. This indicates that general knowledge from base models can also contribute to capability transfer. However, using domain-specific expert models further increases the scores to 61.3, 32.8, and 54.2, outperforming the base model setting by 4.0, 5.2, and 3.6 points, respectively. These results show that the improvements of SuperAI cannot be explained by additional training or general knowledge transfer alone. The specialized knowledge of expert models provides additional benefits for capability fusion. Overall, this experiment confirms that SuperAI can effectively transfer domain-specific knowledge from expert models to heterogeneous host models.

\section{Cross-Layer Matching and Shared Host-Layer Selection}
\label{app:shared_layer_selection}
This section provides additional details on the selection of the shared host layer. Let $\mathcal{L}_h$ denote the candidate host layers and
$\mathcal{L}_s$ denote the candidate layers of source model $s$. For each source model, we use prompts that are shared with the host model:
\begin{equation}
\mathcal{A}_s=\{a_i\}_{i=1}^{n_s}.
\end{equation}
For every candidate layer pair $(\ell_s,\ell_h)$, we extract the hidden state of the last non-padding token from both models. We then apply the normalization, PCA projection, and orthogonal Procrustes alignment defined in Section 3.1 of the main paper. The mapping is fitted on \(A_s^{\mathrm{fit}}\) and evaluated on the disjoint validation set \(V_s\). For a validation prompt \(a_i\), the aligned source coordinate is
\begin{equation}
\hat{z}_{s\to h}^{(\ell_s,\ell_h)}(a_i)
=
\left(R_s^{(\ell_s,\ell_h)}\right)^\top
z_s^{(\ell_s)}(a_i).
\end{equation}
Following the definition of Geometry Match in Section~3.1 of the main paper, we evaluate each candidate layer pair using three complementary quantities. First, paired cosine similarity measures the coordinate-wise agreement between aligned source and host representations:
\begin{equation}
\begin{aligned}
C_{\mathrm{pair},s}(\ell_s,\ell_h)
&=
\frac{1}{|V_s|}
\sum_{a_i\in V_s}
\cos\left(
\hat{z}_{s\to h}^{(\ell_s,\ell_h)}(a_i),
z_h^{(\ell_h,s)}(a_i)
\right).
\end{aligned}
\end{equation}
Second, we compute linear CKA between the source and host PCA coordinate matrices:
\begin{equation}
C_{\mathrm{CKA},s}(\ell_s,\ell_h)
=
\operatorname{CKA}
\left(
Z_s^{\mathrm{val}},
Z_h^{\mathrm{val}}
\right).
\end{equation}
Third, normalized reconstruction error measures the remaining
discrepancy after alignment:
\begin{equation}
E_{\mathrm{nrmse},s}(\ell_s,\ell_h)
=
\frac{
\left\|
\mathbf{Z}^{\mathrm{val}}_s
\mathbf{R}^{(\ell_s,\ell_h)}_s
-
\mathbf{Z}^{\mathrm{val}}_h
\right\|_F
}{
\left\|
\mathbf{Z}^{\mathrm{val}}_h
\right\|_F
}.
\end{equation}
The Geometry Match score used for layer selection is
\begin{equation}
\begin{aligned}
S_s(\ell_s,\ell_h)
= \frac{1}{3}\Bigl[
&\max\!\left(
0,C_{\mathrm{pair},s}(\ell_s,\ell_h)
\right)
\\
&+\operatorname{clip}\!\left(
C_{\mathrm{CKA},s}(\ell_s,\ell_h),0,1
\right)
\\
&+\exp\!\left(
-E_{\mathrm{nrmse},s}(\ell_s,\ell_h)
\right)
\Bigr].
\end{aligned}
\end{equation}
where $\operatorname{clip}(x,0,1)$ truncates $x$ to the interval
$[0,1]$. The three terms measure paired coordinate agreement,
structural similarity, and reconstruction consistency,
respectively. For each candidate host layer, we first select the best-matching
layer from source model $s$:
\begin{equation}
\ell_s^*(\ell_h)
=
\arg\max_{\ell_s\in\mathcal{L}_s}
S_s(\ell_s,\ell_h).
\end{equation}
The shared host layer is then selected by aggregating the best
Geometry Match scores across all $K$ source models:
\begin{equation}
\ell_h^*
=
\arg\max_{\ell_h\in\mathcal{L}_h}
\sum_{s=1}^{K}
S_s\!\left(\ell_s^*(\ell_h),\ell_h\right).
\end{equation}
Finally, the selected layer for source model $s$ is
\begin{equation}
\ell_s^*
=
\ell_s^*(\ell_h^*).
\end{equation}
The low-, medium-, and high-matching
configurations reported in Table~5 of the main paper are obtained
by ranking candidate layer pairs according to the same Geometry
Match score $S_s(\ell_s,\ell_h)$ defined above.

\section{Limitation}
\label{lim}
Our current evaluation focuses on offline benchmark-based capability fusion, where prompts and target tasks are predefined. We have not yet systematically evaluated SuperAI in interactive human–AI settings involving multi-turn dialogue, evolving user intent, or feedback-dependent behavior. While this does not affect the validity of the reported offline fusion results, interactive deployment may introduce additional challenges such as maintaining transferred capabilities consistently across dialogue turns and adapting residual updates to changing conversational context.
Future work will extend SuperAI to interactive settings by incorporating conversation-level representations, turn-aware residual routing, and human-feedback-based evaluation to study how fused capabilities behave under dynamic user interactions.

\section{More Discussion}
\label{dis}

\noindent$\triangleright$ \textbf{\textit{Q1. The method uses ground-truth task labels to train the residual MLP. How can the authors distinguish capability transfer from ordinary supervised adaptation of the host model?}}

We would like to clarify that SuperAI is not equivalent to standard supervised adaptation. The key distinction is that the residual MLP is jointly optimized with both the task loss and the source-derived increment supervision. As shown in Table 4 of the main paper, using only the task loss (\(\lambda_{\text{inc}}=0\)) consistently underperforms the joint objective across all three host models, while using only the increment loss improves representation matching but does not provide the same downstream gains. This suggests that the improvement does not arise from task supervision alone; rather, the source-derived increments provide an additional transfer signal that complements the task objective. The role of the task loss is to adapt the transferred source information to the downstream computation of the host model, which is necessary because representation correspondence alone does not guarantee functional compatibility.

\noindent$\triangleright$ \textbf{\textit{Q2. The improvement over THESEUS and Transport-and-Merge is large. Could this simply be due to the additional trainable residual MLP rather than the proposed cross-model alignment mechanism?}}

Our ablations indicate that the gain cannot be explained by the residual MLP alone. First, removing source-increment supervision and retaining the same task-trained residual module leads to substantially lower performance than the full joint objective. Second, Table 2 of the main paper shows that, under the same downstream training setup, replacing PCA + orthogonal alignment with no alignment, PCA-only mapping, or unconstrained linear mapping consistently degrades performance across the three host architectures. Third, Table 3 of the main paper shows that the nonlinear residual mapper further improves over a linear mapper, indicating that the alignment module and the host-adaptation module play complementary roles rather than one simply substituting for the other. Therefore, the empirical gains result from the combination of cross-model geometric alignment and host-compatible residual adaptation.

\noindent$\triangleright$ \textbf{\textit{Q3. The paper reports very high post-alignment similarity. Could these scores merely reflect overfitting of the Procrustes mapping to paired anchor samples?}}

The reported alignment scores are not computed on the same samples used to fit the mapping. As described in the Appendix, for every candidate source–host layer pair, the normalization, PCA projection, and orthogonal Procrustes mapping are fitted on \(A_s^{\mathrm{fit}}\), while the matching score is evaluated on a disjoint validation set  $V_s$. CKA is included as one component of the Geometry Match score used for layer selection, while Retrieval@1 is reported as an additional diagnostic metric. Moreover, Table 5 of the main paper shows that downstream capability transfer improves systematically as the held-out post-alignment Match score increases from weakly matched to medium and best-matched layer pairs. This provides evidence that the alignment score captures transferable cross-model structure rather than merely fitting the anchor examples.

\noindent$\triangleright$ \textbf{\textit{Q4. Table 1 of the main paper shows that some domains degrade after multi-source fusion. For example, Code decreases from 39.7 to 28.3 on Llama-3.2-3B, from 55.6 to 44.2 on Qwen2.5-3B, and from 11.1 to 1.3 on Pythia-2.8B. How should these results be interpreted?}}

The Code results reflect the trade-off inherent in joint multi-domain fusion rather than the overall effectiveness of capability integration. SuperAI optimizes a single host model across 15 heterogeneous domains, whose objectives and output formats differ substantially. In particular, code generation requires highly structured token-level generation, making its preferred representation updates different from those of many reasoning and classification domains. Consequently, optimizing the shared representation for broad multi-domain performance does not necessarily preserve the optimum of every individual domain.

Importantly, the goal of this setting is joint capability composition rather than independent per-domain specialization, and SuperAI consistently improves the aggregate 15-domain performance across all three hosts. Future work will investigate domain-conditioned weighting or routing to further improve the balance among heterogeneous capabilities.

\end{document}